\documentclass[journal]{IEEEtran}
\usepackage{amsmath,amsfonts}
\usepackage{algorithmic}
\usepackage{algorithm}
\usepackage{array}
\usepackage{textcomp}
\usepackage{stfloats}
\usepackage{url}
\usepackage{verbatim}
\usepackage{graphicx}
\usepackage{cite}

\usepackage{bm}
\usepackage{epstopdf}
\usepackage{color}
\usepackage{subeqnarray}
\usepackage{indentfirst} 
\usepackage{tikz}
\usepackage{multirow}
\usepackage{mathtools}
\usepackage{hyperref}
\usepackage[cal=boondox]{mathalfa}
\usepackage{subcaption}
\usepackage{multirow}
\usepackage[font=footnotesize,skip=5pt]{caption}
\usepackage{todonotes}
\usepackage{siunitx}
\usepackage{comment}
\usepackage{booktabs}

\newcommand{\R}{\mathbb{R}}
\newcommand{\q}{\boldsymbol{q}}
\newcommand{\dq}{\boldsymbol{\dot{q}}}
\newcommand{\ddq}{\boldsymbol{\ddot{q}}}
\newcommand{\taubf}{\boldsymbol{\tau}}
\newcommand{\g}{\boldsymbol{g}}

\newcommand{\x}{\boldsymbol{x}}
\newcommand{\y}{\boldsymbol{y}}
\newcommand{\norm}[1]{\lVert#1\rVert} 

\newcommand{\kernel}{{\textit{Lagrangian Inspired Polynomial kernel}}}
\newcommand{\kernelInitials}{LIP}

\newtheorem{prop}{Proposition}
\newtheorem{remark}{Remark}

\begin{document}

\title{A Black-Box Physics-Informed Estimator based on Gaussian Process Regression  for Robot Inverse Dynamics Identification}

\author{Giulio~Giacomuzzo,~\IEEEmembership{Member,~IEEE,}  Ruggero~Carli,~\IEEEmembership{Member,~IEEE,}  Diego~Romeres,~\IEEEmembership{Member,~IEEE,} Alberto~Dalla~Libera,~\IEEEmembership{Member,~IEEE}
\thanks{Giulio Giacomuzzo, Ruggero Carli and Alberto Dalla Libera are with the Department of Information Engineering, University of Padova, 35131 Padova, Italy (email: giacomuzzo@dei.unipd.it; carlirug@dei.unipd.it; dallaliber@dei.unipd.it)}
\thanks{Diego Romeres is with the Mitsubishi Electric Research Lab (MERL), Cambridge, MA 02139 USA (email: romeres@merl.com)}
}

\maketitle

\begin{abstract}
Learning the inverse dynamics of robots directly from data, adopting a black-box approach, is interesting for several real-world scenarios where limited knowledge about the system is available. In this paper, we propose a black-box model based on Gaussian Process (GP) Regression for the identification of the inverse dynamics of robotic manipulators. The proposed model relies on a novel multidimensional kernel, called \textit{Lagrangian Inspired Polynomial} (\kernelInitials{}) kernel. The \kernelInitials{} kernel is based on two main ideas. First, instead of directly modeling the inverse dynamics components, we model as GPs the kinetic and potential energy of the system. The GP prior on the inverse dynamics components is derived from those on the energies by applying the properties of GPs under linear operators. Second, as regards the energy prior definition, we prove a polynomial structure of the kinetic and potential energy, and we derive a polynomial kernel that encodes this property. 
As a consequence, the proposed model allows also to estimate the kinetic and potential energy without requiring any label on these quantities.
Results on simulation and on two real robotic manipulators, namely a 7 DOF Franka Emika Panda, and a 6 DOF MELFA RV4FL, show that the proposed model outperforms state-of-the-art black-box estimators based both on Gaussian Processes and Neural Networks in terms of accuracy, generality and data efficiency. The experiments on the MELFA robot also demonstrate that our approach achieves performance comparable to fine-tuned model-based estimators, despite requiring less prior information.
\end{abstract}

\section{Introduction}\label{sec:intro}
Robot manipulators are one of the most widespread platforms both in industrial and service robotics. In many applications involving such systems, control performance strongly benefits from the presence of accurate dynamics models. Inverse dynamics models, which express joint torques as a function of joint positions, velocities and accelerations, are fundamental in different control problems, ranging from high-precision trajectory tracking
\cite{khosla1988experimental, siciliano}
to detection and estimation of contact forces \cite{ADL_propriocept19, haddadin2017robot, de2008atlas}. 

Despite their importance, the derivation of accurate inverse dynamics models is still a challenging task and several techniques have been proposed in the literature. Traditional model-based approaches derive parametric models directly from first principles of physics, see, for instance, \cite{hollerbach2008model,panda-id,sousa2014physical, Kinodynamic-ID}. Their performance, however, is often limited by both the presence of parameter uncertainty and the inability to describe certain complex dynamics typical of real systems, such as motor friction or joint elasticity. 

For this reason, in recent years there has been an increased interest in deriving inverse dynamics models by means of machine learning. Several data-driven techniques have been proposed, mainly based on deep neural networks (NN) \cite{Goodfellow-et-al-2016} and Gaussian Process Regression (GPR) \cite{rasmussen2003gaussian}. In this context, both gray-box and black-box approaches have been explored. Within gray-box techniques, a model-based component encoding the known dynamics is combined with a data-driven one, which compensates for modeling errors and unknown dynamical effects, \cite{romeres2016online,Camoirano,Nguyen-prior-model,Romeres-TAC}.
However, advantages of these methods strongly depend on the effectiveness of the model-based component, so they still require accurate physical models, whose derivation might be particularly time-consuming and complex.

In contrast, pure black-box methods learn inverse dynamics models directly from experimental data, without requiring deep knowledge of the underlying physical system. Despite their ability to approximate even complex non-linear dynamics, pure black-box methods typically suffer from low data efficiency and poor generalization properties: learned models require a large amount of samples to be trained and extrapolate only within a neighborhood of the training trajectories.  

Several solutions were proposed to overcome the aforementioned limitations, see, for instance, \cite{Rueckert-LSTM, Polydoros-real-time-NN} in the context of NN, and \cite{Sparse_GP_RT_control, gijsberts2011incremental,nguyen2009model,Rezaei-cascaded-gp} for the GPR framework. A promising class of them is represented by \textit{Physics Informed Learning} (PIL) \cite{karniadakis2021physics}, which proposes to embed insights from physics as a prior in black-box models\cite{ADL_GIP19, lutter2019_delan, cheng-vector-valued-RKHS-4invDyn, evangelisi-physically-consistent, yoon2024kinematics, lee2023robot}. Instead of learning the inverse dynamics in an unstructured manner, which makes the problem unnecessarily hard, physical properties are embedded in the model to improve generalization and data efficiency.

In this manuscript, we propose a PIL model for inverse dynamics identification of mechanical systems based on GPR. 
When applying GPR to the inverse dynamics identification, the standard approach consists of modeling directly each joint torque with a distinct Gaussian Process (GP), assuming the GPs independent of one another given the current joint position, velocity, and acceleration. This strategy, hereafter denoted as the single-output approach, simplifies the regression problem but ignores the correlations between the different joint torques imposed by the Lagrangian equations, which in turn could limit generalization and data efficiency.

In contrast, we propose a multi-output GPR estimator based on a novel kernel function, named \kernel{} (\kernelInitials{}), which exploits Lagrangian mechanics to model also the correlations between the different joint torques.
Our method is based on two main ideas: first, instead of modeling directly joint torques, we model as GPs the kinetic and potential energy of the system. Driven by the fact that the dynamics equations are linear w.r.t. the Lagrangian, we obtain the torques GPs by applying a set of linear operators to the GPs of the potential and kinetic energy. Second, as regards the prior definition, we show that the kinetic and potential energy are polynomial functions in a suitable input space, and we derive a polynomial kernel that encodes this property.

The collected results show that the \kernelInitials{} estimator outperforms state-of-the-art black-box GP estimators as well as NN-based solutions, obtaining better data efficiency and generalization performance. This fact confirms that encoding physical properties in the models is a promising strategy to improve data efficiency. Interestingly, experiments carried out on the MELFA robot demonstrate that our approach achieves out-of-sample estimation performance more than comparable to fine-tuned model-based estimators, despite requiring less prior information. Finally, we validated the effectiveness of the energy estimation both in simulation and on the Franka Emika Panda.

\textbf{Related work:}
The idea of encoding Lagrangian equations in black-box models has already been explored in literature. For instance, recently a NN model has been introduced in \cite{lutter2019_delan}, named Deep Lagrangian Networks (DeLan). The DeLan model adopts two distinct feedforward NN: one for modeling the inertia matrix elements and another for the potential energy. Based on these estimates, the torques are subsequently computed by implementing Lagrangian equations. Among kernel-based methods, \cite{cheng-vector-valued-RKHS-4invDyn} formulated inverse dynamics identification as an optimization problem in a Reproducing Kernel Hilbert Space (RKHS). They derived a multi-output kernel by modeling the entire Lagrangian with a standard Square Exponential (SE) kernel. Consequently, this formulation lacks the capability to estimate kinetic and potential energies. Furthermore, both the experiments in \cite{cheng-vector-valued-RKHS-4invDyn} and our own work carried out using the model in \cite{cheng-vector-valued-RKHS-4invDyn} demonstrate out-of-sample performance comparable to that of single-output models. In the GPR domain, \cite{evangelisi-physically-consistent} defined a GP prior both on the potential energy and on the elements of the inertia matrix using the standard SE kernel, leading to a high number of hyperparameters to be optimized. Moreover, the efficacy of the approach in \cite{evangelisi-physically-consistent} has only been validated on a simulated 2-DOF example.

The approaches in \cite{lutter2019_delan,cheng-vector-valued-RKHS-4invDyn,evangelisi-physically-consistent} try to improve data efficiency by imposing the structure of the Lagrangian equations, without considering specific basis functions tailored to the inverse dynamics, such as polynomial functions. The fact that the inverse dynamics spans a polynomial space has already been observed in \cite{Ge1998_struct} and further explored in \cite{ADL_GIP19} and \cite{Cheng-structured-RKHS-4invDyn}. In \cite{Ge1998_struct}, the authors proposed a NN to model the elements of the inertia matrix and the gravity term as polynomials. The approach has been validated on a simulated 2 DOF example and generalization to more complex systems has not been investigated. At the increase of the DOF, the number of NN parameters necessary to model the inertia matrix increases exponentially, which could make the training of the network particularly difficult.
The kernel-based formulation provides a straightforward solution to encode polynomials structure. Both in \cite{ADL_GIP19} and \cite{Cheng-structured-RKHS-4invDyn}, the authors adopt estimators based on polynomial kernels to model each joint torque, without considering output correlations. Results confirm that performance improves with respect to standard black-box solutions. To the best of our knowledge, none of the approaches presented in the literature proposes a multi-output inverse dynamics estimator that models the kinetic and potential energies using tailored polynomial kernels.

Our contribution is twofold. First, we prove the polynomial structure of the kinetic and potential energy and we derive the \kernelInitials{} estimator, a black-box multi-output GPR model that encodes the symmetries typical of Lagrangian systems. Second, we show that, differently from single-output GP models, the \kernelInitials{} model we propose can estimate the kinetic and potential energy in a principled way, allowing its integration with energy-based control strategies \cite{lutter2019_delan_control}. 
We compare the \kernelInitials{} model performance against baselines and state-of-the-art algorithms through extensive tests on simulated setups of increasing complexity and also two real manipulators, a Franka Emika Panda and a Mitsubishi MELFA robot. Furthermore, we carried out a trajectory tracking experiment on the Franka Emika Panda robot to test the effectiveness of our estimation for control.

The paper is organized as follows.
Section \ref{sec:background} reviews the theory of GPR for inverse dynamics identification. In Section \ref{sec:proposed_approach}, we present the proposed approach. First, we show how to derive the GP prior on the torques from the one on the energies, exploiting the laws of Lagrangian mechanics; then we describe the polynomial kernel we use to model the system energies; finally, we present the energy estimation algorithm. In Section \ref{sec:related-works} we provide an overview of the main related works presented in the literature.
Section \ref{sec:experiments} reports the performed experiments, while Section \ref{sec:conclusions} concludes the paper.

\section{Background}\label{sec:background}
In this section, we describe the inverse dynamics identification problem, and we concisely review the GPR framework for multi-output models. In addition, we introduce some notions and properties (polynomial kernels; application of linear operators to GPs) that will play a fundamental role in the derivation of the novel estimation scheme we propose.

\subsection{Inverse dynamics}
Consider an $n$-degrees of freedom (DOF) serial manipulator composed of $n+1$ links connected by $n$ joints, labeled $1$ through $n$. Let $\boldsymbol{q}_t =\left[q_t^1,\ldots, q_t^n\right]^T \in \mathbb{R}^n$ and $\boldsymbol{\tau}_t=\left[\tau_t^1,\ldots,\tau_t^n\right]^T \in \mathbb{R}^n$ be the vectors collecting, respectively, the joint coordinates and generalized torques at time $t$, where $q_t^i$ and $\tau_t^i$ denote, respectively, the joint coordinate and the generalized torque of joint $i$. 
Moreover, $\dot{\boldsymbol{q}}_t$ and $\ddot{\boldsymbol{q}}_t$ denote, respectively, the joint velocity and acceleration vectors. 
For ease of notation, in the following we will denote explicitly the dependence on $t$ only when strictly necessary.

The inverse dynamics identification problem consists of identifying the map that relates $\boldsymbol{q}_t, \dot{\boldsymbol{q}}_t, \ddot{\boldsymbol{q}}_t$ with $\boldsymbol{\tau}_t$, given a dataset of input-output measures $\mathcal{D}$. Under rigid body assumptions, the dynamics equations derived from first principle of physics, for instance by applying Lagrangian or Hamiltonian mechanics, are described by the following matrix equation
\begin{equation}\label{eq:dyn_eq}
    B(\boldsymbol{q}) \ddot{\boldsymbol{q}} + \boldsymbol{c}(\boldsymbol{q}, \dot{\boldsymbol{q}}) + \boldsymbol{g}(\boldsymbol{q})  + \bm{\varepsilon} = \boldsymbol{\tau} \text{,}
\end{equation}
where $B(\boldsymbol{q})$ is the inertia matrix, $\boldsymbol{c}(\boldsymbol{q}, \dot{\boldsymbol{q}})$ and $\boldsymbol{g}(\boldsymbol{q})$ account for the contributions of fictitious forces and gravity, respectively, and $\bm{\varepsilon}$ is the torque due to friction and unknown dynamical effects. We refer the interested reader to \cite{siciliano} for a complete and detailed description and derivation of \eqref{eq:dyn_eq}. 
In general, the terms in \eqref{eq:dyn_eq} are nonlinear w.r.t. $\boldsymbol{q}_t$ and $\dot{\boldsymbol{q}}_t$, and depend on two important sets of parameters, that is, the kinematics and dynamics parameters, hereafter denoted by $\boldsymbol{w}_k$ and $\boldsymbol{w}_d$, respectively. It is worth stressing that the vector $\boldsymbol{w}_k$ depends on the convention adopted to derive the kinematic relations. For instance, a possible choice is given by the Denavit-Hartenberg convention, see\cite{denavit1955kinematic}. Instead, $\boldsymbol{w}_d$ is a vector including the mass, the position of the center of mass, the elements of the inertia tensor, and the friction coefficients of each link.  
Typically, the vector $\boldsymbol{w}_k$ is known with high accuracy, while tolerances on $\boldsymbol{w}_d$ are much more consistent. Discrepancies between the nominal and actual values of $\boldsymbol{w}_d$ can be so considerable that \eqref{eq:dyn_eq} with nominal parameters is highly inaccurate and unusable for advanced control applications.

\subsection{Gaussian Process Regression for multi-output models}\label{subsec:GP}

A relevant class of solutions proposed for inverse dynamics identification relies on GPR. GPR is a principled probabilistic framework for regression problems that allows estimating an unknown function given a dataset of input-output observations. Let $f:\mathbb{R}^m \rightarrow \mathbb{R}^d$ be the unknown function and let $\mathcal{D}=\{X, Y\}$ be the input-output dataset, composed of the input set $X=\{\boldsymbol{x}_1 \dots \boldsymbol{x}_N\}$ and the output set $Y=\{\boldsymbol{y}_1,\dots, \boldsymbol{y}_N\}$, 
where $\x_i \in \mathbb{R}^m$ and $\boldsymbol{y}_i \in \mathbb{R}^d$, $i=1, \dots, N$.  
We assume the following measurement model
\begin{equation}\label{eq:GP-generative-model-componentwise}
\boldsymbol{y}_i = f(\boldsymbol{x}_i) + \boldsymbol{e}_i, \qquad i=1,\ldots,N,
\end{equation}
where $\boldsymbol{e}_i$ is a zero-mean Gaussian noise with variance $\Sigma_{e_i} \in \mathbb{R}^d \times \mathbb{R}^d$, i.e.,  $\boldsymbol{e}_i \sim \mathcal{N}(0, \Sigma_{e_i})$, independent from the unknown function. We assume that $\Sigma_{e_i}$ is a diagonal matrix, that is,
$$
\Sigma_{e_i} = \text{diag}(\sigma_{e_1}^2, \dots, \sigma_{e_d}^2),
$$
where $\sigma_{e_j}^2$ denotes the variance of the noise affecting the $j$-th component of $f$. By letting $\boldsymbol{y}=\left[\boldsymbol{y}_1^T,\dots, \boldsymbol{y}_N^T\right]^T$ and $\boldsymbol{e}=\left[\boldsymbol{e}_1^T,\dots, \boldsymbol{e}_N^T\right]^T$ we can write
\begin{equation}\label{eq:GP-generative-model}
    \boldsymbol{y} = \begin{bmatrix} \boldsymbol{y}_1 \\ \vdots \\ \boldsymbol{y}_N \end{bmatrix} = \begin{bmatrix} f(\x_1) \\ \vdots \\ f(\x_N) \end{bmatrix} + \begin{bmatrix} \boldsymbol{e}_1 \\ \vdots \\ \boldsymbol{e}_N \end{bmatrix} = f(X)+ \boldsymbol{e},
\end{equation}
where the noises $\boldsymbol{e}_1,\dots, \boldsymbol{e}_N$ are assumed independent and identically distributed. It turns out that the variance of $\boldsymbol{e}$ is a block diagonal matrix with equal diagonal blocks, namely
$$
\Sigma_e = \text{diag}(\Sigma_{e_1}, \dots, \Sigma_{e_N}),
$$
with $\Sigma_{e_1}=\Sigma_{e_2}=\ldots =\Sigma_{e_N}$.

The unknown function $f$ is defined a priori as a GP, that is, $f \sim GP(m(\x), k(\x, \x'))$, where $m(\cdot) : \mathbb{R}^m \rightarrow \mathbb{R}^d$ is the prior mean and $k(\cdot, \cdot) : \mathbb{R}^m \times \mathbb{R}^m \rightarrow \mathbb{R}^{d \times d}$ is the prior covariance, also termed kernel. The function $k(\cdot, \cdot)$, which typically depends on a set of hyperparameters $\boldsymbol{\theta}$, represents the covariance between the values of the unknown function in different input locations, that is, $\text{Cov}[f(\boldsymbol{x}_{p}),f(\boldsymbol{x}_{q})] = k(\x_{p}, \x_q)$. 
As an example, in the scalar case ($d=1$), a common choice for $k(\cdot, \cdot)$ is the Square Exponential (SE) kernel, which defines the covariance between samples based on the distance between their input locations. More formally
\begin{equation} \label{eq:K-SE}
    k_{SE}(\x, \x') = \lambda e^{-\norm{\x - \x'}^2_{\Sigma^{-1}}},
\end{equation}
where $\lambda$ and $\Sigma$ are the kernel hyperparameters.

Under the Gaussian assumption, the posterior distribution of $f$ given $\mathcal{D}$ in a general input location $\boldsymbol{x}$ is still a Gaussian distribution, with mean and variance given by the following expressions
\begin{subequations} \label{eq:GP-posterior}
\begin{align}
    &\mathbb{E}[f(\boldsymbol{x})|\mathcal{D}] = m(\x) + K_{\boldsymbol{x}X}(K_{XX}+ \Sigma_e)^{-1}(\boldsymbol{y}-\boldsymbol{m}_X),\label{eq:gp-mean}\\
    &\text{Cov}[f(\boldsymbol{x})|\mathcal{D}] = k(\x, \x) -K_{\boldsymbol{x}X}(K_{XX} + \Sigma_e)^{-1} K_{X\boldsymbol{x}},  \label{eq:gp-var}
\end{align}
\end{subequations}
where $K_{\boldsymbol{x}X} \in \mathbb{R}^{d \times dN}$ is given by
\begin{equation}
    K_{\boldsymbol{x}X} = K_{X\boldsymbol{x}}^T = \begin{bmatrix}
        k(\x, \x_1), \dots, k(\x, \x_N)
    \end{bmatrix},
\end{equation}
and $K_{XX} \in \mathbb{R}^{dN \times dN}$ is the block matrix
\begin{equation}
    \label{eq:cov_matrix}
    K_{XX} = \begin{bmatrix} k(\x_1, \x_1) & \dots & k(\x_1, \x_N) \\
             \vdots & \ddots & \vdots \\
             k(\x_N, \x_1) & \dots & k(\x_N,\x_N)\end{bmatrix}.
\end{equation}
See \cite{rasmussen2003gaussian} for a detailed derivation of formulas in \eqref{eq:GP-posterior}. The posterior mean \eqref{eq:gp-mean} is used as an estimate of $f$, that is, $\hat{f}=\mathbb{E}[f(\boldsymbol{x})|\mathcal{D}]$, while \eqref{eq:gp-var} is useful to derive confidence intervals of $\hat{f}$.

In the remainder of this section we review additional notions and properties of GPR, that will be fundamental in deriving the estimator proposed in this paper.
 \subsubsection{Polynomial kernel} \label{sec:poly-ker} The \kernelInitials{} model relies on the use of standard polynomial kernels (see Eq. \eqref{eq:poly} below). As discussed in \cite{rasmussen2003gaussian} and \cite{Scholkopf_LWK}, by exploiting the Reproducing Kernel Hilbert Space (RKHS) interpretation of GPR, a polynomial kernel constrains $f(\boldsymbol{x})$ to belong to the space of polynomial functions in the components of $\boldsymbol{x}$. Specifically, the standard polynomial kernel of degree $p$, expressed as
\begin{equation}\label{eq:poly}
    k^{(p)}_{pk}(\x, \x') = (\x^T\Sigma_{pk}\x' + \sigma_{pk})^p
\end{equation}
defines the space of inhomogeneous polynomials of degree $p$ in the elements of $\boldsymbol{x}$, that is, the space generated by all the monomials in the elements of $\boldsymbol{x}$ with degree $r$, $0 \leq r \leq p$. The matrix $\Sigma_{pk}$ and the scalar value $\sigma_{pk}$ are hyperparameters that determine the weights assigned to the different monomials. Typically, $\Sigma_{pk}$ is assumed to be diagonal. By setting $\sigma_{pk}=0$ we obtain the so-called homogeneous polynomial kernel of degree $p$, hereafter denoted by $k^{(p)}_{hpk}(\x, \x')$, which identifies the space generated by all the monomials in the elements of $\boldsymbol{x}$ with degree $p$. 

For future convenience, we introduce a notation to point out polynomial functions in a compact way. We denote by $\mathbb{P}_{(p)}(\x_{(p_1)})$  the set of polynomial functions of degree not greater than $p$ defined over the elements of $\x$, such that each element of $\x$ appears with degree not greater than $p_1$. For instance, the polynomial kernel in \eqref{eq:poly} identifies $\mathbb{P}_{(p)}(\x_{(p)})$. The notation naturally extends to the multi-input case: $\mathbb{P}_{(p)}(\x_{(p_x)}, \boldsymbol{z}_{(p_z)})$ denotes the set of polynomial functions of degree not greater than $p$ defined over the elements of $\x$ and $\boldsymbol{z}$, where the elements of $\x$ (resp. $\boldsymbol{z}$) appears with degree not greater than $p_x$ (resp. $p_z$).

\subsubsection{Combination of kernels} \label{sec:poly-comb}Kernels can be combined through sum or multiplication \cite{rasmussen2003gaussian,Scholkopf_LWK}. 
Let $f_1(\x)$ and $f_2(\x)$ be two Gaussian processes and let $k_1(\x,\x')$ and $k_2(\x,\x')$ be the kernels associated to $f_1$ and $f_2$, respectively. Then both $f_s(\x)=f_1(\x)+f_2(\x)$ and $f_p(\x)=f_1(\x)f_2(\x)$ are Gaussian processes; if $k_s(\x,\x')$ and $k_p(\x,\x')$ are the kernels associated to $f_s(\x)$ and $f_p(\x)$, respectively, then $k_s(\x,\x') = k_1(\x,\x')+k_2(\x,\x')$ and $k_p(\x,\x') = k_1(\x,\x')\,k_2(\x,\x')$.
Now, assume that both $k_1$ and $k_2$ are polynomial kernels and let $\mathcal{M}_1$ and $\mathcal{M}_2$ be the sets of monomials generating the polynomial spaces spanned by $k_1$ and $k_2$, respectively. Then, both $k_s$ and $k_p$ are polynomial and 
\begin{itemize}
\item $k_s$ spans the polynomial space generated by the set $\mathcal{M}_s = \mathcal{M}_1 \cup \mathcal{M}_2$;
\item $k_p$ spans the polynomial space generated by the set $\mathcal{M}_p$ composed by all the monomials obtained as the product of one monomial of $\mathcal{M}_1$ with one monomial of $\mathcal{M}_2$. 
\end{itemize}

\subsubsection{Linear operators and GPs} \label{sec:linOP-GP} 
Assume now that $f \sim GP(m_f(\x), k_f(\x, \x^\prime))$ is a scalar Gaussian process, that is $f:\mathbb{R}^m \to \mathbb{R}$. Let $\mathcal{G}$ be a linear operator on the realizations of $f$. 
We assume that the operator produces functions with range in $\mathbb{R}^r$, defined on the same domain of the argument, namely $g = \mathcal{G}f : \mathbb{R}^m \rightarrow \mathbb{R}^r$. 
In this setup, the operator $\mathcal{G}$ has $r$ components, $\mathcal{G}_1, \ldots, \mathcal{G}_r$, where $\mathcal{G}_i$ maps $f$ into the $i$-th component of $g$, that is 
$$
g = \mathcal{G}f =[\mathcal{G}_1 f\,\, \dots \,\,\mathcal{G}_r f]^T.
$$ 

As GPs are closed under linear operators (see \cite{sarkka2011linear}), $g$ is still a GP, i.e. $g \sim GP(m_g(\x), k_g(\x, \x^\prime))$. Its mean and covariance are given by applying $\mathcal{G}$ to the mean and covariance of the argument $f$, resulting in 
\begin{equation}
    m_g(\x) = \mathbb{E}\left[\mathcal{G} f(\x)\right] = \mathcal{G} m_f(\x) : \mathbb{R}^m \to \mathbb{R}^r
\end{equation}
and
\begin{align}\label{eq:lin_op_cov}
    &k_g(\x, \x^\prime) = \text{Cov} \left[\mathcal{G} f(\boldsymbol{x}),\mathcal{G} f(\boldsymbol{x}^\prime) \right]  : \mathbb{R}^m \times \mathbb{R}^m \to \mathbb{R}^{r \times r} \\
    &k_g(\x, \x^\prime) =\begin{bmatrix}
    \mathcal{G}_1 \mathcal{G}_1^\prime k_f(\boldsymbol{x}, \boldsymbol{x}^\prime)  &  \dots & \mathcal{G}_1 \mathcal{G}_r^\prime k_f(\boldsymbol{x}, \boldsymbol{x}^\prime)  \\
    \vdots & \ddots & \vdots\\
    \mathcal{G}_r \mathcal{G}_1^\prime k_f(\boldsymbol{x}, \boldsymbol{x}^\prime)  &  \dots & \mathcal{G}_r \mathcal{G}_r^\prime k_f(\boldsymbol{x}, \boldsymbol{x}^\prime)
    \end{bmatrix}.
\end{align}
where $\mathcal{G}_j^\prime$ is the same operator as $\mathcal{G}_j$ but applied to $k_f(\boldsymbol{x}, \boldsymbol{x}^\prime)$ as function of $\x^\prime$. In details, the notation $\mathcal{G}_i \mathcal{G}_j^\prime k_f(\boldsymbol{x}, \boldsymbol{x}^\prime)$ means that $\mathcal{G}_j^\prime$ is first applied to $k_f(\boldsymbol{x}, \boldsymbol{x}^\prime)$ assuming $\boldsymbol{x}$ constant and then $\mathcal{G}_i$ is applied to the obtained result assuming $\boldsymbol{x}^\prime$ constant. For convenience of notation, we denote $k_g(\x, \x^\prime)$ by $\mathcal{G} \mathcal{G}^\prime k_f(\boldsymbol{x}, \boldsymbol{x}^\prime)$. Finally, the cross-covariance between $f$ and $\mathcal{G}f$ at input locations $\x$ and $\x^\prime$ is given as
\begin{subequations}\label{eq:cross-covariances_linop}
\begin{align}
\text{Cov}[\mathcal{G} f(\x), f(\x')] &=\mathcal{G} k_f(\x,\x'), \\\label{eq:cross-covariances_linop_b}
\text{Cov}[f(\x), \mathcal{G}f(\x')] &=\left[\mathcal{G}' k_f(\x,\x')\right]^T,
\end{align}
\end{subequations}
with
\begin{subequations}
\begin{align}\label{eq:cross-covariances_linop2}
\text{Cov}[\mathcal{G} f(\x), f(\x')]& =\left[\mathcal{G}_1 k_f(\x,\x') \ldots \mathcal{G}_r k_f(\x,\x')\right]^T, \\
\text{Cov}[f(\x), \mathcal{G}f(\x')] &=\left[\mathcal{G}'_1 k_f(\x,\x') \ldots \mathcal{G}'_r k_f(\x,\x')\right],
\end{align}
\end{subequations}
where again the notation $\mathcal{G} k_f$ and $\mathcal{G}' k_f$ is used to indicate when the operator acts on $k_f$ as function of $\x$ and $\x'$, respectively.
We refer the interested reader to \cite{sarkka2011linear} for a detailed discussion on GPs and linear operators.

\subsection{GPR for Inverse Dynamics Identification}
\label{subsec:GPR_ID}

When GPR is applied to inverse dynamics identification, the inverse dynamics map is treated as an unknown function and modeled a priori as a GP. The GP-input at time $t$ is $\boldsymbol{x}_t = (\boldsymbol{q}_t, \dot{\boldsymbol{q}}_t, \ddot{\boldsymbol{q}}_t)$, while outputs are torques. The standard approach consists in defining the GP prior directly on the inverse dynamics function, by assuming its $n$ components to be conditionally independent given the GP input $\boldsymbol{x}_t$. 
As a consequence, the overall inverse dynamics identification problem is split into a set of $n$ scalar and independent GPR problems,
$$
y_t^i = f^i(\x_t) + e_t^i
$$
where the $i$-th torque component $f^i : \mathbb{R}^{3n} \rightarrow \mathbb{R}$ is estimated independently of the others as in \eqref{eq:GP-posterior} with $d=1$ and $\boldsymbol{y} = \boldsymbol{y}^i = [y_1^i \dots y_N^i]^T$ , being $y_t^i$ a measure of the $i$-th torque at time $t$.

Observe that the conditionally independence assumption is an approximation of the actual model, and it might limit generalization and data efficiency. As described in the next section, we propose a multi-output GP model that naturally correlates the different torque dimensions, thus obtaining the following generative model,
$$
\y_t = f(\x_t) + \boldsymbol{e}_t,
$$
with $f : \mathbb{R}^{3 n} \rightarrow \mathbb{R}^{n}$, where $n$ is the number of DOF of the considered mechanical system, $\boldsymbol{y}_t\ \in \mathbb{R}^n$ is the vector of torque measurements at time $t$ and $\boldsymbol{e}_t\ \in \mathbb{R}^n$ is the noise at time $t$ modeled as in Sec.~\ref{subsec:GP}. The estimate of $f$ can now be computed as described in Sec.~\ref{subsec:GP}, with $d=n$.

\section{Lagrangian Inspired Polynomial Model}
\label{sec:proposed_approach}

In this section, we derive the proposed multi-output GP model for inverse dynamics learning, named \textit{Lagrangian Inspired Polynomial} (\kernelInitials{}) model. 
In traditional GP-based approaches, a GP prior is directly defined on the joint torques. In the \kernelInitials{} framework, instead, we model 
 the kinetic and potential energies as two different GPs and derive the GP prior of the torques exploiting the laws of Lagrangian mechanics and the properties reviewed in Section \ref{sec:linOP-GP}. The section is organized as follows. First, in Section~\ref{subsec:derivation}, the inverse dynamics GP-models are derived from the GPs of the kinetic and potential energies, without specifying the structure of the involved kernels. The novel polynomial priors assigned to the energies are introduced and discussed in Section~\ref{subsec:prior_functions}. Finally, an algorithm to estimate the system energies is proposed in Section \ref{sec:energy_est}.

\subsection{From energies to torques GP models} \label{subsec:derivation}
Let $\mathcal{T}(\boldsymbol{q}, \dot{\boldsymbol{q}})$ and $\mathcal{V}(\boldsymbol{q})$ be, respectively, the kinetic and potential energy of a $n$-DOF system of the form \eqref{eq:dyn_eq}. The \kernelInitials{} model assumes that $\mathcal{T}(\boldsymbol{q}, \dot{\boldsymbol{q}})$ and $\mathcal{V}(\boldsymbol{q})$ are two independent zero-mean GPs with covariances determined by the kernel functions $k^{\mathcal{T}}(\boldsymbol{x}, \boldsymbol{x}^\prime)$ and $k^\mathcal{V}(\boldsymbol{x}, \boldsymbol{x}^\prime)$, that is 
\begin{subequations}\label{eq:energy-prior}
\begin{align}
    \mathcal{T} \sim GP(0, k^{\mathcal{T}}(\boldsymbol{x}, \boldsymbol{x}^\prime)) \label{eq:pot-prior},\\
    \mathcal{V} \sim GP(0, k^{\mathcal{V}}(\boldsymbol{x}, \boldsymbol{x}^\prime)) \label{eq:kin-prior}.
\end{align}
\end{subequations}
where $\boldsymbol{x} = (\boldsymbol{q}, \dot{\boldsymbol{q}}, \ddot{\boldsymbol{q}})$ as defined in Section~\ref{subsec:GPR_ID}. The GP prior on $\mathcal{T}$ and $\mathcal{V}$ cannot be used directly in GPR to compute posterior distributions since kinetic and potential energies are not measured. However, starting from the prior on the two energies, we can derive a GP prior for the torques by relying on Lagrangian mechanics. Lagrangian mechanics states that the inverse dynamics equations in \eqref{eq:dyn_eq} (with $\bm{\varepsilon}=0$), also named Lagrange's equations, are the solution of a set of differential equations of the Lagrangian function $\mathcal{L} = \mathcal{T}(\boldsymbol{q}, \dot{\boldsymbol{q}}) - \mathcal{V}(\boldsymbol{q})$ \cite{siciliano}. The $i$-th differential equation of \eqref{eq:dyn_eq} is
\begin{equation}\label{eq:lagrangian-eq}
    \frac{d\mathcal{L}}{dt}\left(\frac{\partial \mathcal{L}}{\partial \dot{q}^i}\right) - \frac{\partial \mathcal{L}}{\partial q^i}  = \tau^i,
\end{equation}
where $q^i$, $\dot{q}^i$, and $\tau^i$ are, respectively, the $i$-th component of $\boldsymbol{q}$, $\dot{\boldsymbol{q}}$, and $\boldsymbol{\tau}$. Equation \eqref{eq:lagrangian-eq} involves an explicit differentiation w.r.t. time that can be avoided using the chain rule, leading to
the following linear partial differential equation of $\mathcal{L}$, 
\begin{equation}\label{eq:lagrangian-eq-chain}
    \sum_{j = 1}^n  \left(\frac{\partial^2 \mathcal{L}}{\partial \dot{q}^i\partial \dot{q}^j}\ddot{q}^j + \frac{\partial^2 \mathcal{L}}{\partial \dot{q}^i\partial q^j}\dot{q}^j\right)  - \frac{\partial \mathcal{L}}{\partial q^i}= \tau^i. 
\end{equation}
Now it is convenient to introduce the linear operator $\mathcal{G}_i$ that maps $\mathcal{L}$ in the left-hand side of \eqref{eq:lagrangian-eq-chain}:
\begin{equation*}
    \mathcal{G}_i \mathcal{L} = \sum_{j = 1}^n  \left(\frac{\partial^2 \mathcal{L}}{\partial \dot{q}^i\partial \dot{q}^j}\ddot{q}^j + \frac{\partial^2 \mathcal{L}}{\partial \dot{q}^i\partial q^j}\dot{q}^j\right)  - \frac{\partial \mathcal{L}}{\partial q^i}= \tau^i,
\end{equation*}
In compact form we can write
\begin{equation}\label{eq:lagrangian-operator}
    \boldsymbol{\tau} = \mathcal{G}\mathcal{L} = [\mathcal{G}_1 \mathcal{L}\,\, \dots \,\,\mathcal{G}_n \mathcal{L}]^T.
\end{equation}
where the above equality defines the linear operator $\mathcal{G}$ mapping $\mathcal{L}$ into $\boldsymbol{\tau}$.

Notice that, in the \kernelInitials{} model, $\mathcal{L}$ is a GP since $\mathcal{T}$ and $\mathcal{V}$ are two independent GPs. Indeed, as pointed out in Sec.~\ref{sec:poly-comb}, the sum of two independent GPs is a GP, and its kernel is the sum of the kernels, namely,
\begin{subequations}\label{eq:L}
\begin{align}
    &\mathcal{L}\sim GP(0, k^\mathcal{L}(\x,\x^\prime)),\label{eq:L-GP}\\
    &k^\mathcal{L}(\boldsymbol{x}, \boldsymbol{x}^\prime) = k^\mathcal{T}(\boldsymbol{x}, \boldsymbol{x}^\prime) + k^\mathcal{V}(\boldsymbol{x}, \boldsymbol{x}^\prime).\label{eq:L-kernel}
\end{align}
\end{subequations}

Equation \eqref{eq:lagrangian-operator} shows that, in our setup, the inverse dynamics map is the result of the application of the linear operator $\mathcal{G}$ to the GP defined in \eqref{eq:L-GP} and \eqref{eq:L-kernel} modeling $\mathcal{L}$. 
Based on the properties discussed in Section \ref{sec:linOP-GP}, we can conclude that the inverse dynamics is modeled as a zero mean GP with covariance function obtained by applying \eqref{eq:lin_op_cov}. Specifically, we have that 
 $\boldsymbol{\tau} \sim GP(0,k^{\tau}(\boldsymbol{x}, \boldsymbol{x}^\prime))$ where
\begin{equation}\label{eq:lagrangian-kernel}
    k^{\tau}(\boldsymbol{x}, \boldsymbol{x}^\prime)= 
    \begin{bmatrix}
    \mathcal{G}_1 \mathcal{G}_1^\prime k^\mathcal{L}(\boldsymbol{x}, \boldsymbol{x}^\prime)  &  \dots & \mathcal{G}_1 \mathcal{G}_n^\prime k^\mathcal{L}(\boldsymbol{x}, \boldsymbol{x}^\prime)  \\
    \vdots & \ddots & \vdots\\
    \mathcal{G}_n \mathcal{G}_1^\prime k^\mathcal{L}(\boldsymbol{x}, \boldsymbol{x}^\prime)  &  \dots & \mathcal{G}_n \mathcal{G}_n^\prime k^\mathcal{L}(\boldsymbol{x}, \boldsymbol{x}^\prime) 
    \end{bmatrix}. 
\end{equation}
It is worth stressing that the proposed approach models the inverse dynamics function as an unknown multi-output function $f(\boldsymbol{x}): \mathbb{R}^{3 n} \rightarrow \mathbb{R}^{n}$. 
The training of the LIP model consists of computing the posterior distribution of $f$ given $\mathcal{D}$ at a general input location $\boldsymbol{x}$ as expressed by \eqref{eq:GP-posterior} and described in Sec.~\ref{subsec:GPR_ID}; the inputs are position, velocities and accelerations of the joints, while outputs are torques.

\subsection{Kinetic and potential energy polynomial priors}
\label{subsec:prior_functions}
In this section, we derive the kernel functions $k^\mathcal{V}$ and $k^\mathcal{T}$ adopted in the \kernelInitials{} model to define the priors on the potential and kinetic energies. 

The definition of $k^\mathcal{V}$ and $k^\mathcal{T}$ rely on the existence of two suitable transformations mapping positions and velocities of the generalized coordinates
into two sets of variables with respect to which the kinetic energy and potential energy are polynomial functions.

We start our analysis by introducing the main elements of the aforementioned transformations. Let $\q^i \text{ and }\dq^i$ be the vectors containing the positions and velocities of the joints up to index $i$, respectively, i.e.,
\begin{eqnarray*}
		&&\q^i = 
		\left[ q^1 \,,\, \dots \,,\,q^i \right]^T \in \mathbb{R}^{i}
		\text{,} \\
		&&\dq^i = 
		\left[\dot{q}^1 \,,\, \dots \,,\, \dot{q}^i \right]^T \in \mathbb{R}^{i} \text{.}
  \end{eqnarray*}
  Now, assume the manipulator to be composed by $N_r$ revolute joints and $N_p$ prismatic joints, with $N_r$ and $N_p$ such that $N_r + N_p = n$. Then, let $I_r=\{r_{1}, \dots, r_{N_r}\}$, $r_1<r_2<\ldots < r_{N_r}$, and $I_p=\{p_1, \dots, p_{N_p}\}$, $p_1<p_2<\ldots < p_{N_p}$, be the sets containing the revolute and prismatic joints indexes, respectively. Accordingly, let us introduce the vectors
\begin{eqnarray*}
		&&\boldsymbol{q}_{c} = 
		\left[ \cos\left(q^{r_1}\right) \,,\, \dots \,,\, \cos\left(q^{r_{N_r}}\right) \right]^T \in \mathbb{R}^{N_r}
		\text{,} \\
		&&\boldsymbol{q}_{s} = 
		\left[\sin\left(q^{r_1}\right) \,,\, \dots \,,\, \sin\left(q^{r_{N_r}}\right) \right]^T \in \mathbb{R}^{N_r} \text{,}\\
		&&\boldsymbol{q}_{p} = 
		\left[ q^{p_1} \,,\, \dots \,,\, q^{p_{N_p}} \right]^T \in \mathbb{R}^{N_p}\text{.}
	\end{eqnarray*}
By $q_{c}^b$, $q_{s}^b$ and $q_{p}^b$ we denote the $b$-th element of $\q_c$,  $\q_s$ and $\q_{p}$, respectively. 

Next, let $I_r^i$ (resp. $I_p^i$) be the subset of $I_r$ (resp. $I_p$) composed by the indexes lower or equal to $i$, and let us define the vectors $\q^{i}_{c}$, $\q^{i}_{s}$, (resp. $\q_{p}^i$) as the restriction of $\q_c$, $\q_s$ (resp. $\q_p$) to $I_r^i$ (resp. $I_p^i$). 
For the sake of clarity, consider the following example. Let index $i$ be such that $r_j \leq i < r_{j+1}$ for some $1\leq j < r_{N_r}$. Then $I_r^i= 
		\left\{ r_1 \,,\, \dots \,,\,r_j  \right\} \in \mathbb{R}^{j}$, $\boldsymbol{q}_{c}^i = 
		\left[ \cos\left(q^{r_1}\right) \,,\, \dots \,,\, \cos\left(q^{r_j}\right) \right]^T \in \mathbb{R}^{j}$ and $\boldsymbol{q}_{s}^i = 
		\left[\sin\left(q^{r_1}\right) \,,\, \dots \,,\, \sin\left(q^{r_j}\right) \right]^T \in \mathbb{R}^{j}$.

To conclude, let $\q_{cs_b}$ be the vector concatenating the $b$-th elements of $\q_c$ and $\q_s$, that is, $\q_{cs_b} = [q_{c}^b \, , q_{s}^b]^T$.

Next, we continue our analysis by considering first the design of $k^\mathcal{V}$ and then the design of $k^\mathcal{T}$.

\subsubsection{Potential energy}
The following proposition establishes that the potential energy is polynomial w.r.t. the set of variables $(\q_c, \q_s, \q_p)$ that, as previously highlighted, are functions of the joint positions vector $\q$.
\vspace{5pt}
\begin{prop} \label{prop:pot_pol}
Consider a manipulator with $n+1$ links and $n$ joints. The total potential energy $\mathcal{V}(\boldsymbol{q})$ belongs to the space $\mathbb{P}_{(n)}(\q_{c_{(1)}}, \q_{s_{(1)}}, \q_{p_{(1)}})$, namely it is a polynomial function in $(\q_c, \q_s, \q_p)$ of degree not greater than $n$, such that each element of $\q_c$, $\q_s$ and $\q_p$ appears with degree not greater than $1$.
Moreover, for any monomial of the aforementioned polynomial, the sum of the degrees of $q_{c}^b$ and $q_{s}^b$ is equal or lower than \num{1}, namely, it holds
\begin{equation}\label{eq:cos-sin-U-constr}
    deg(q_{c}^b) + deg(q_{s}^b) \leq 1. 
\end{equation}
\end{prop}
\vspace{5pt}
The proof follows from similar concepts exploited in \cite{ADL_GIP19} and is reported in the Appendix.

To comply with the constraints on the maximum degree
of each term, we adopt a kernel function given by the product
of $N_r + N_p$ inhomogeneous kernels of the type defined in equation \eqref{eq:poly}, where
\begin{itemize}
\item $N_r$ kernels have $p = 1$ and each of them is defined on
the $2$-dimensional input space given by $\q_{cs_b}$, $b\in I_r$;
\item $N_p$ kernels have $p = 1$ and each of them is defined on the
$1$-dimensional input $q_p^b$, $b\in I_p$.
\end{itemize}
The resulting kernel is then given by
\begin{equation}\label{eq:potential-kernel}
        k^\mathcal{V}(\x, \x') = \prod_{b \in I_r}k^{(1)}_{pk}(\q_{cs_b}, \q'_{cs_b}) \prod_{b \in I_p} k^{(1)}_{pk}(q_p^b, q_p^{\prime b}).
\end{equation}
Few observations are now needed. Firstly notice that each of the $n$ kernels accounts for the contribution of a distinct joint. Secondly, exploiting properties reviewed in \ref{sec:poly-comb}, one can easily see  that kernel in \eqref{eq:potential-kernel} spans $\mathbb{P}_{(n)}(\q_{c_{(1)}}, \q_{s_{(1)}}, \q_{p_{(1)}})$. Finally, since all the $N_r$ kernels defined on $\q_{cs_b}$, $b\in I_r$, have $p=1$, also the constraint in 
\eqref{eq:cos-sin-U-constr} is satisfied.

\subsubsection{Kinetic energy}
We start by observing that the total kinetic energy is the sum of the kinetic energies relative to each link, that is,
\begin{equation}\label{eq:tot_kin}
    \mathcal{T}(\boldsymbol{q}, \dot{\boldsymbol{q}}) = \sum_{i=1}^{n} \mathcal{T}_i(\boldsymbol{q}, \dot{\boldsymbol{q}}),
\end{equation}
where $\mathcal{T}_i(\boldsymbol{q}, \dot{\boldsymbol{q}})$ is the kinetic energy of Link $i$.
The following proposition establishes that $\mathcal{T}_i$ is polynomial w.r.t. the set of variables $(\q^i_c, \q^i_s, \q^i_p, \dq^{i})$, which are functions of the joints positions and velocities vectors $\q^i$ and $\dq^i$.
\vspace{5pt}
\begin{prop} \label{prop:kin_pol}
 Consider a manipulator with $n+1$ links and $n$ joints. The kinetic energy $\mathcal{T}_i(\q, \dq)$ of link $i$ belongs to $\mathbb{P}_{(2i+2)}(\q^{i}_{c_{(2)}}, \q^{i}_{s_{(2)}}, \q^{i}_{p_{(2)}} \dq^{i}_{(2)})$, namely it is a polynomial function in $(\q^i_c, \q^i_s, \q^i_p, \dq^{i})$ of degree not greater than $2i+2$, such that : 
 \begin{itemize}
 \item[(i)] each element of $\q^{i}_{c}$, $\q^{i}_{s}$, $\q_{p}^i$ and $\dq^{i}$ appears with degree not greater than $2$;
 \item[(ii)] each monomial has inside a term of the type $\dot{q}^i \dot{q}^j$ for $1\leq i \leq n$ and $i\leq j \leq n$; and 
 \item[(iii)] in any monomial the sum of the degrees of $q_{c}^b$ and $q_{s}^b$ is equal or lower than \num{2}, namely
  \begin{equation}\label{eq:cos-sin-T-constr}
      deg(q_{c}^b) + deg(q_{s}^b) \leq 2. 
  \end{equation}
 \end{itemize}
\end{prop}
\vspace{5pt}
The proof follows from similar concepts exploited in \cite{ADL_GIP19} and is reported in the Appendix.

To comply with the constraints and properties stated in the above Proposition, we adopt a kernel function given by the product
of $i$ inhomogeneous kernels of the type defined in equation \eqref{eq:poly}, and $\num{1}$ homogeneous kernel, where
\begin{itemize}
\item $|I_r^i|$ inhomogeneous kernels have $p = 2$ and each of them is defined on
the $2$-dimensional input space given by $\q_{cs_b}$, $b\in I_r$;
\item $|I_p^i|$ inhomogeneous kernels have $p = 2$ and each of them is defined on the
$1$-dimensional input $q_p^b$, $b\in I_p$;
\item $1$ homogeneous kernel has $p=2$ and is defined on the $i$-dimensional input $\dq^i$.
\end{itemize}
The resulting kernel is then given by
\begin{equation} \label{eq:kinetic_LGI}
\begin{aligned}
    k^\mathcal{T}_i(\x, \x') = \,&k_{hpk}^{(2)}(\dq^i, \dq'^{i}) \, \cdot\\
    &\prod_{b \in I_r^i} k_{pk}^{(2)}(\q_{{cs}_b}, \q'_{{cs}_b}) \, \cdot 
    \prod_{b \in I_p^i} k_{pk}^{(2)}(q_p^b, q_p^{\prime b}).
\end{aligned}
\end{equation}
Also in this case some observations are needed. Firstly notice that 
$|I_r^i| + |I_p^i|=i$. Secondly, the fact that all the kernels have $p=2$ ensures that properties in $(i)$ and $(iii)$ of the above Proposition are satisfied. Thirdly, using an homogeneous kernel defined on $\dq^i$ with $p=2$ guarantees the validity of property $(ii)$.

Based on \eqref{eq:tot_kin} and on the properties reviewed in Section \ref{sec:poly-comb}, we finally define $k^\mathcal{T}$ as
\begin{equation}\label{eq:total_kin_k}
    k^\mathcal{T}(\x, \x') = \sum_{i=1}^n k^\mathcal{T}_i(\x, \x').
\end{equation}

To conclude, in this paper we introduce the multi-output torque prior $k^\tau$ expressed as in \eqref{eq:lagrangian-kernel}, where $k^{\mathcal{L}}$ is defined in \eqref{eq:L-kernel} adopting for 
$k^\mathcal{V}$ and $k^\mathcal{T}$ the polynomial structures in 
\eqref{eq:potential-kernel} and \eqref{eq:total_kin_k}, respectively. The kernel $k^\tau$ built in this way is termed the LIP kernel.

\begin{remark}
Notice that Proposition \ref{prop:pot_pol} characterizes the potential energy $\mathcal{V}$ of the whole system, while Proposition \ref{prop:kin_pol} focuses on the kinetic energy of each link. 
In general, characterizing the energies for each link and designing a tailored kernel to be combined as done for $k^\mathcal{T}$ should allow for higher flexibility in terms of regularization and for more accurate predictions. We have experimentally verified this fact for the kinetic energy, while no significant advantages have been obtained for the potential energy. This is the reason why for the latter we have decided to adopt a more compact description involving the whole system. 
\end{remark}

\subsection{Energy estimation}\label{sec:energy_est}
The proposed \kernelInitials{} model provides a principled way to estimate the kinetic and potential energy from the torque measurements $\y$ \footnote{We note that the Lagrangian satisfying \eqref{eq:lagrangian-eq} is not unique. For example, the potential energy could be arbitrarily shifted by a constant. As the energies are estimated from the torque measures, the information on such a shift is lost and the estimation could be affected by a constant offset, which needs to be taken into account in applications such as energy tracking.} 
Indeed, in our model, $\mathcal{T}$, $\mathcal{V}$ and $\boldsymbol{\tau}$ are jointly Gaussian distributed, since the prior of  $\taubf$ is derived by applying the linear operator $\mathcal{G}$ to the kinetic and potential GPs $\mathcal{T}$ and $\mathcal{V}$. The covariances between $\mathcal{T}$ and $\taubf$ and between $\mathcal{V}$ and $\taubf$ at general input locations $\x$ and $\x'$ are
\begin{subequations}
\begin{align}
    &\text{Cov}[\mathcal{T}(\x), \taubf(\x^\prime)] = \text{Cov}[\mathcal{T}(\x), \mathcal{G}\mathcal{L}(\x^\prime)] = k^{\mathcal{T}\tau}(\x, \x'),\\
    &\text{Cov}[\mathcal{V}(\x), \taubf(\x^\prime)] = \text{Cov}[\mathcal{V}(\x), \mathcal{G}\mathcal{L}(\x^\prime)] = k^{\mathcal{V}\tau}(\x, \x').
\end{align}
\end{subequations}
In view of \eqref{eq:cross-covariances_linop} and recalling that we model $\mathcal{T}$ and $\mathcal{V}$ as independent GPs, we obtain
\begin{subequations}
    \begin{equation}
        \begin{split}
            k^{\mathcal{T}\tau}(\x, \x') &= \text{Cov}[\mathcal{T}(\x), \mathcal{G}\mathcal{T}(\x^\prime) - \mathcal{G}\mathcal{V}(\x^\prime)]\\
            &= \text{Cov}[\mathcal{T}(\x), \mathcal{G}\mathcal{T}(\x^\prime)] \\ &=\left[\mathcal{G}^\prime k^{\mathcal{T}}(\x^\prime)\right]^T \\
            &= [\mathcal{G}'_1 k^\mathcal{T}(\x, \x'), \, \dots \, , \mathcal{G}_n'k^\mathcal{T}(\x, \x')]
        \end{split}
    \end{equation}
    and
    \begin{equation}
        \begin{split}
           k^{\mathcal{V}\tau}(\x, \x') &= \text{Cov}[\mathcal{V}(\x), \mathcal{G}\mathcal{T}(\x^\prime) - \mathcal{G}\mathcal{V}(\x^\prime)]\\
            &= \text{Cov}[\mathcal{V}(\x), -\mathcal{G}\mathcal{V}(\x^\prime)] \\ 
            &= -\left[\mathcal{G}^\prime k^{\mathcal{V}}(\x, \x^\prime)\right]^T\\
            &= -[\mathcal{G}'_1 k^\mathcal{V}(\x, \x'), \, \dots \, , \mathcal{G}_n'k^\mathcal{V}(\x, \x')].
        \end{split}
    \end{equation}
\end{subequations}

The Gaussian property makes the posterior distributions of $\mathcal{T}$ and $\mathcal{V}$ given $\mathcal{D}$ known and analytically tractable. At any general input location $\x$, these posteriors are Gaussians with means
\begin{subequations}
\begin{align}
    &E[\mathcal{T}(\boldsymbol{x})|\mathcal{D}] = K^{\mathcal{T}\tau}_{\boldsymbol{x}X}(K_{XX}+ \Sigma_e)^{-1}\y,\\
    &E[\mathcal{V}(\boldsymbol{x})|\mathcal{D}] = K^{\mathcal{V}\tau}_{\boldsymbol{x}X}(K_{XX}+ \Sigma_e)^{-1}\y
\end{align}
\end{subequations}
and variances
\begin{subequations}
\begin{align}
    &\mathbb{V}[\mathcal{T}(\x)] = k^\mathcal{T}(\x, \x) - K^{\mathcal{T}\tau}_{\x X}(K_{XX} + \Sigma_e)^{-1} (K^{\mathcal{T}\tau}_{\x X})^T,\\
    &\mathbb{V}[\mathcal{V}(\x)] = k^\mathcal{V}(\x, \x) - K^{\mathcal{V}\tau}_{\x X}(K_{XX} + \Sigma_e)^{-1} (K^{\mathcal{V}\tau}_{\x X})^T
\end{align}
\end{subequations}
where the covariance matrices $K^{\mathcal{T}\tau}_{\x X} \in \mathbb{R}^{1\times nN}$ and $K^{\mathcal{V}\tau}_{\x X} \in \mathbb{R}^{1\times nN}$ are obtained as
\begin{subequations}
\begin{align}
    &K^{\mathcal{T}\tau}_{\x X} = \begin{bmatrix} k^{\mathcal{T} \tau}(\x, \x_1),\dots, k^{\mathcal{T} \tau}(\x, \x_N)\end{bmatrix},\\
    &K^{\mathcal{V}\tau}_{\x X} = \begin{bmatrix} k^{\mathcal{V} \tau}(\x, \x_1),\dots, k^{\mathcal{V} \tau}(\x, \x_N)\end{bmatrix},
\end{align} 
\end{subequations}
while $K_{XX}$ is computed using \eqref{eq:cov_matrix} as described in Sec.~\ref{subsec:GPR_ID}.
\section{Discussion on related works} \label{sec:related-works}
 Several algorithms have been developed in the literature to identify the inverse dynamics. The different solutions can be grouped according to if and how they exploit prior information on the model and, more specifically, on the nominal kinematic and dynamic parameters, i.e., $\boldsymbol{w}_k$ and $\boldsymbol{w}_d$, respectively. 

An important class of solutions assumes to have an accurate a-priori knowledge of $\boldsymbol{w}_k$ and reformulate the inverse dynamics identification problem in terms of estimation of $\boldsymbol{w}_d$, see, for instance, \cite{hollerbach2008model,panda-id,sousa2014physical}. These approaches are based on the fact that \eqref{eq:dyn_eq} is linear w.r.t. $\boldsymbol{w}_d$, namely,
\begin{equation}\label{eq:linear_model}
    \boldsymbol{\tau} = \Phi(\boldsymbol{q}, \dot{\boldsymbol{q}}, \ddot{\boldsymbol{q}}) \boldsymbol{w}_d
\end{equation}
where $\Phi(\boldsymbol{q}, \dot{\boldsymbol{q}}, \ddot{\boldsymbol{q}}) \in \mathbb{R}^{n \times n_d}$ is a matrix of nonlinear functions derived from \eqref{eq:L} assuming $\boldsymbol{w}_k$ known, see \cite{siciliano}; $n_d$ is the dimension of $\boldsymbol{w}_d$. Then, the dynamics parameters $\boldsymbol{w}_d$ are identified by solving the least squares problem associated to the set of noisy measurements $\boldsymbol{y} = \Phi(X) \boldsymbol{w}_d +\boldsymbol{e}$, where $\Phi(X) \in \mathbb{R}^{(n\cdot N)\times n_d}$ is the matrix obtained stacking the $\Phi$ matrices evaluated in the training inputs. The accuracy of these approaches is closely related to the accuracy of $\Phi(\boldsymbol{q}, \dot{\boldsymbol{q}}, \ddot{\boldsymbol{q}})$ and, in turn, of $\boldsymbol{w}_k$. 
To compensate for uncertainties on the  knowledge of $\boldsymbol{w}_k$, the authors in \cite{Kinodynamic-ID} have proposed a kinodynamic identification algorithm that identifies simultaneously $\boldsymbol{w}_k$ and $\boldsymbol{w}_d$. Though being an interesting and promising approach, this algorithm requires additional sensors to acquire the link poses, which are not always available. 

It is worth stressing that, in several cases, deriving accurate models of $\Phi(\boldsymbol{q}, \dot{\boldsymbol{q}}, \ddot{\boldsymbol{q}})$ is a quite difficult and time-consuming task, see \cite{siciliano}. This fact has motivated the recent increasing interest in gray-box and black-box solutions. In gray-box approaches,  the model is typically given by the sum of two components. The first component is based on a first principles model of \eqref{eq:dyn_eq}, for instance, $\boldsymbol{\tau} = \Phi(\x) \hat{\boldsymbol{w}}_d$, where $\hat{\boldsymbol{w}}_d$ is an estimate of $\boldsymbol{w}_d$. The second one is a black-box model that compensates for model inaccuracies by exploiting experimental data. Several GP-based solutions have been proposed in this setup,\cite{Nguyen-prior-model,Camoirano,romeres2016online,Romeres-TAC}. Generally, GPR algorithms model the torque components as $n$ independent GPs and include the model-based component either in the mean or in the covariance functions. In the first case, the prior mean of the $i$-th GP is $\boldsymbol{\phi}^i(\boldsymbol{x}) \hat{\boldsymbol{w}}_d$, where $\boldsymbol{\phi}^i(\boldsymbol{x})$ is the $i$-th row of $\Phi(\boldsymbol{x})$, and the covariance is modeled by a SE kernel of the form in \eqref{eq:K-SE}, see, for instance, \cite{Nguyen-prior-model}. Instead, when the prior model is included in the covariance function, the prior mean is assumed null in all the configurations  and the covariance is given by the so-called semiparametric kernels, see \cite{Camoirano,romeres2016online,Nguyen-prior-model,Romeres-TAC}, expressed as
\begin{equation}\label{eq:K-SP}
    K_{SP}(\boldsymbol{x}, \boldsymbol{x}^\prime) = \boldsymbol{\phi}^i(\boldsymbol{x}) \Gamma \boldsymbol{\phi}^i(\boldsymbol{x}^\prime)^T + K_{SE}(\boldsymbol{x}, \boldsymbol{x}^\prime).
\end{equation}
Observe that the first term in the right-hand-side of \eqref{eq:K-SP} is a linear kernel in the features $\boldsymbol{\phi}^i(\boldsymbol{x})$, where the elements of the matrix $\Gamma \in \mathbb{R}^{n_d \times n_d}$ are the hyperparameters. We refer the interested reader to \cite{Nguyen-prior-model} for an exhaustive discussion on the two aforementioned strategies. The advantages of gray-box solutions w.r.t. black-box models depend on the accuracy of the model-based prior and the generalization properties of the black-box model. If the black-box correction generalizes also to unseen input locations, such models can provide an accurate estimate of the inverse dynamics.

To avoid the presence of model bias or other model inaccuracies or to deal with the unavailability of first principle models, several works proposed in the literature have adopted pure black-box solutions; this is the case of also the approach developed in this work. The main criticality of these solutions is related to generalization properties: when the number of DOF increases, the straightforward application of a black-box model to learn the inverse dynamics provides accurate estimates only in a neighborhood of the training input locations. In the context of NN, the authors in \cite{Polydoros-real-time-NN} have tried to improve generalization by resorting to a recurrent neural network architecture, while in \cite{Rueckert-LSTM} an LSTM network has beeen adopted. 

As regards GP-based solutions, several algorithms have introduced GP approximations to derive models trainable with sufficiently rich datasets. Just to mention some of the proposed solutions, \cite{Sparse_GP_RT_control} has adopted the use of sparse GPs, \cite{gijsberts2011incremental} has approximated the kernel with random features, \cite{nguyen2009model} has relied on the use of local GPs rather than of a single GP model, \cite{Rezaei_cascaded_GP} has proposed an approach inspired by the Newton-Euler algorithm to reduce dimensionality and improve data efficiency. Interestingly, to improve generalization, the works in \cite{Cheng-structured-RKHS-4invDyn} and \cite{ADL_GIP19} have introduced novel kernels specifically tailored to inverse dynamics identification rather than adopting standard SE kernels of the form in \eqref{eq:K-SE}; it is known that the latter kernels define a regression problem in an infinite dimensional Reproducing Kernel Hilbert Space (RKHS) and this could limit generalization properties. For instance, following ideas introduced in \cite{Ge1998_struct}, the authors in \cite{ADL_GIP19} have proposed a polynomial kernel named Geometrically Inspired Polynomial kernel (GIP) associated to a finite-dimensional RKHS that contains the rigid body dynamics equations. Experiments carried out in \cite{ADL_GIP19} and \cite{Cheng-structured-RKHS-4invDyn} show that limiting the RKHS dimensions can significantly improve generalization and data efficiency.

It is worth remarking that most of the works proposed in the context of GPR, model the joint torques as a set of independent single-output GPs. To overcome the limitations raising from neglecting the existing correlations, the authors in \cite{Williams-multi-task-GP} have applied a general multitask GP model for learning the robot inverse dynamics. The resulting model outperforms the standard single-output model equipped with SE kernel, in particular when samples of each joint torque are collected in different portions of the input space, which, however, is a rather rare case in practice. 
A recent research line aims at deriving multi-output models in a black-box fashion by incorporating in the priors the correlations inherently forced by \eqref{eq:lagrangian-eq}, as proposed in this work. To the best of our knowledge, within the framework of kernel-based estimators, the first example has been presented in \cite{cheng-vector-valued-RKHS-4invDyn}. Differently from our approach, in \cite{cheng-vector-valued-RKHS-4invDyn} the authors model the RKHS of the entire Lagrangian function by an SE kernel, instead of modeling separately the kinetic and potential energies using specific polynomial kernels. As a consequence, the algorithm does not allow estimating kinetic and potential energies, that can be useful for control purposes. Experiments carried out in \cite{cheng-vector-valued-RKHS-4invDyn} show out-of-sample performance similar to the one of single-output models. A recent solution within the GPR framework, developed independently and in parallel to this work, has been presented in \cite{evangelisi-physically-consistent}. In this work the authors have defined GP priors, based on standard SE kernels, on the potential energy and on the elements of the inertia and stiffness matrices, thus leading to a high number of hyperparameters to be optimized. In \cite{evangelisi-physically-consistent}, the feasibility and effectiveness of the proposed approach has been validated only in a simulated 2-DOF example. 

The Lagrangian equations \eqref{eq:lagrangian-eq} have inspired also several NN models, see, for instance, the model recently introduced in \cite{lutter2019_delan}, named Deep Lagrangian Networks (DeLan). In DeLan two distinct feedforward NN have been adopted to model on one side the inertia matrix elements and on the other side the potential energy; then the torques are computed by applying \eqref{eq:lagrangian-eq}. Experiments carried out in \cite{lutter2019_delan} have focused more on tracking control performance than on estimation accuracy, showing the effectiveness of Delan for tracking control applications. Interestingly, the same authors in \cite{lutter2019_delan_control} have carried out experiments in two physical platforms, a cart-pole and a Furuta pendulum, showing that kinetic and potential energies estimated by DeLan can be used to derive energy-based controllers. As highlighted in \cite{cranmer2019lagrangian}, the main criticality of these models is parameters initialization: indeed, due to the highly nonlinear objective function to be optimized, starting from random initializations the Stochastic Gradient Descent (SGD) algorithm quite often gets stuck in local minima. For this reason, in order to identify a set of parameters exhibiting satisfactory performance, it is necessary to perform the network training multiple times starting from different initial conditions.

\section{Experiments}\label{sec:experiments}
In order to evaluate the performance of the \kernelInitials{} model, we performed extended experiments both on simulated and on real setups. All the estimators presented in this section have been implemented in Python and using the functionalities provided by the library PyTorch \cite{pytorch}. The code is publicly available \footnote{\url{https://github.com/merlresearch/LIP4RobotInverseDynamics}}.

\subsection{Simulation experiments}\label{sec:exp_sim} 

\begin{figure*}
     \centering
     \includegraphics[width=\textwidth]{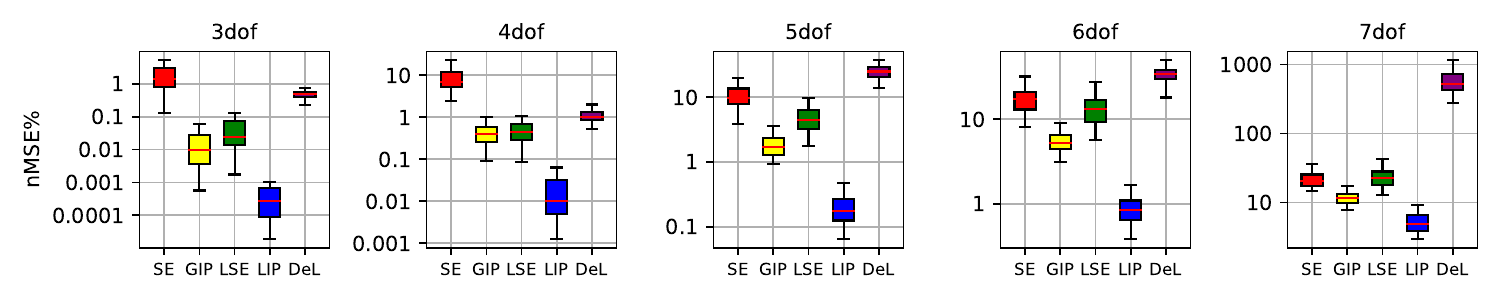}
     \caption{Box plots of the average torque nMSE obtained with the simulations described in Section \ref{sec:gen}. 
     }
     \label{fig:cum_nmse}
\end{figure*}

\begin{figure*}
     \centering
     \begin{subfigure}[b]{0.69\textwidth}
        \centering
        \includegraphics[width=\textwidth]{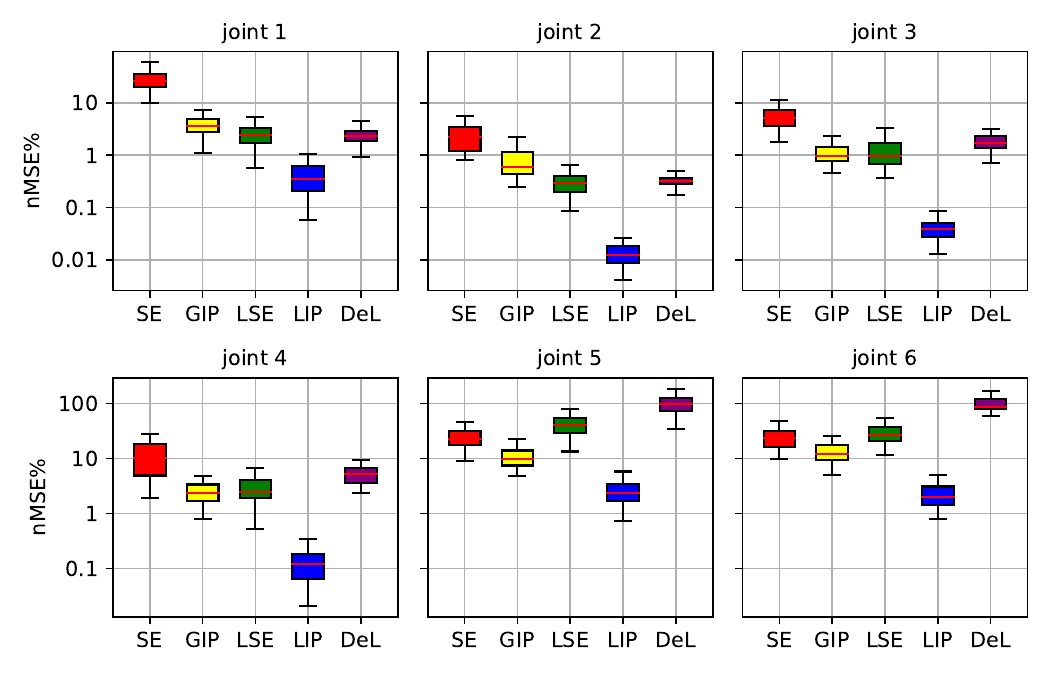}
        \caption[with table]{}
        \vspace{5pt}
        \begin{footnotesize}
            \begin{tabular}{ l c c c c c c}
                \hline \\[-1.5ex]
        
                    & J1 & J2 & J3 & J4 & J5 & J6 \\ [0.5ex]
        
                \hline \\ [-1.5ex]
        
                SE    & 26.40
                    & 2.19
                    & 5.13
                    & 10.33
                    & 22.80
                    & 23.68
                 \\ [0.5ex]
                GIP    & 3.58
                    & 0.59
                    & 0.98
                    & 2.40
                    & 9.86
                    & 11.93
                 \\ [0.5ex]
                LSE    & 2.47
                    & 0.30
                    & 0.95
                    & 2.47
                    & 40.19
                    & 27.16
                 \\ [0.5ex]
                LIP    & \textbf{0.35}
                    & \textbf{0.01}
                    & \textbf{0.04}
                    & \textbf{0.12}
                    & \textbf{2.34}
                    & \textbf{2.02}
                 \\ [0.5ex]
                DeL    & 2.31
                    & 0.32
                    & 1.74
                    & 5.20
                    & 97.47
                    & 86.68
                 \\ [0.5ex]
                \hline
            \end{tabular}
        \end{footnotesize}
        \vspace{5pt}
        \label{fig:gen}
     \end{subfigure}
     \hfill
     \begin{subfigure}[b]{0.295\textwidth}
         \centering
         \includegraphics[width=\textwidth]{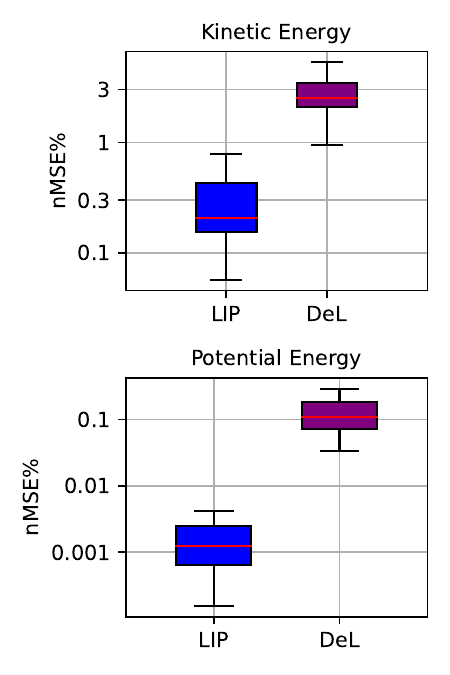}
         \caption{}
\vspace{22pt}
\begin{footnotesize}
        \begin{tabular}{ l c c}
        \hline \\[-1.5ex]
            & Kinetic & Potential \\ [0.5ex]
        \hline \\ [-1.5ex]
        LIP & \textbf{0.21} & \textbf{0.00} \\ [0.5ex]
        DeL & 2.50 & 0.11 \\ [0.5ex]

        \hline
    \end{tabular}
\end{footnotesize}
\vspace{22pt}
         \label{fig:gen_energy}
     \end{subfigure}
     \hfill
\caption{Box plots of the torque nMSE (a) and enregy nMSE (b) obtained with the simulations described in Section \ref{sec:gen}. 
The tables below the figures report the numerical values of the median nMSE percentages.}
\end{figure*}

The first tests are performed on a simulated Franka Emika Panda robot, which is a 7 DOF manipulator with only revolute joints. The joint torques of the robot are computed directly from desired joint trajectories through the inverse dynamics equations in \eqref{eq:dyn_eq}, with the common assumption $\bm{\varepsilon}=0$, i.e., there are no unknown components in the dynamics. The dynamics equations have been implemented using the Python package \textit{Sympybotics}\footnote{\url{https://github.com/cdsousa/SymPyBotics}},

\subsubsection{Generalization} \label{sec:gen}
In the first set of experiments, we tested the ability of the estimators to extrapolate on unseen input locations. To obtain statistically relevant results, we designed a Monte Carlo (MC) analysis composed of \num{50} experiments. Each experiment consists in collecting two trajectories, one for training and one for test. Both the training and test trajectories are sum of $N_s=50$ random sinusoids. For each realization, the trajectory of the $i$-th joint is
\begin{equation}
    q_i(t) = \sum_{l=1}^{N_s} \frac{a}{\omega_f\,l}\,\sin(\omega_f\, l \, t) - \frac{b}{\omega_f\,l}\,\cos(\omega_f\, l\, t), 
    \label{eq:trajectories_gen}
\end{equation}

\begin{table}[h!]
    \centering
    \begin{tabular}{|c|c|c|c|c|c|c|c|c|}
        \hline
          &J1 & J2 & J3 & J4 & J5 & J6 & J7& unit\\ 
         \hline
         $\q_{min}$  & \footnotesize{-2.3} &\footnotesize{-1.4} &\footnotesize{-2.3} &\footnotesize{-2.7} & \footnotesize{-2.3} &\footnotesize{0.2} & \footnotesize{-2.3} & rad\\
         \hline
         $\q_{max}$  & \footnotesize{2.3} &\footnotesize{1.4} &\footnotesize{2.3} &\footnotesize{-0.3} & \footnotesize{2.3} &\footnotesize{3.2} & \footnotesize{2.3} & rad \\
         \hline
         $\dq_{max}$ & \footnotesize{2.2} & \footnotesize{2.2} & \footnotesize{2.2} & \footnotesize{2.2} & \footnotesize{2.6} & \footnotesize{2.6} & \footnotesize{2.6} & rad/s\\
         \hline 
         $\ddq_{max}$& \footnotesize{15} & \footnotesize{7.5} & \footnotesize{10} & \footnotesize{12.5} & \footnotesize{15} & \footnotesize{20} & \footnotesize{20} & $\text{rad/s}^2$\\
         \hline
       
         \hline
    \end{tabular}
    \caption{Limits on joint position, velocity, and acceleration considered when generating the trajectories of the Monte-Carlo experiment described in section \ref{sec:gen}. Velocity ranges are $\dot{q_i} \in [-\dot{q_i}_{max}, \dot{q_i}_{max}]$, while acceleration ranges are $\ddot{q_i} \in [-\ddot{q_i}_{max}, \ddot{q_i}_{max}]$.}
    \label{tab:joint_limits}
\end{table}

with $\omega_f = \SI[per-mode=symbol]{0.02}{\radian \per \second}$, while $a$ and $b$ are sampled from a uniform distribution ranging in $[-c,\,\, c]$, with $c$ chosen in order to respect the limits on joint position, velocity and acceleration, which are reported in Table~\ref{tab:joint_limits}. A zero-mean Gaussian noise with standard deviation \SI{0.01}{\newton\meter} was added to the torques of the training dataset to simulate measurement noise. All the generated datasets are composed of 500 samples, collected at a frequency of \SI{10}{\hertz}. 

We compared the \kernelInitials{} model with both GP-based estimators and a neural network-based estimator. Three different GP-based baselines are considered. Two of them are single-output models, one based on the Square Exponential (SE) kernel in \eqref{eq:K-SE} and the other one based on the GIP kernel presented in \cite{ADL_GIP19}. The third solution, instead, is the multi-output model presented in \cite{cheng-vector-valued-RKHS-4invDyn}, hereafter denoted as LSE. The LSE estimator models directly the Lagrangian function, instead of modeling the kinetic and potential energy separately. The Lagrangian is modeled using a SE kernel defined on an augmented input space obtained by substituting the positions of the revolute joints with their sine and cosine. The authors did not release the code to the public and thus we made our own implementation of the method.
The hyperparameters of all the GP-based estimators have been optimized by marginal likelihood maximization \cite{rasmussen2003gaussian}. 

The neural network baseline is a DeLan network. To select the network architecture, we performed a grid search experiment comparing different configurations, and we chose the one providing the best perfomance on the test trajectories. The network has been trained with standard SGD optimization techniques. Regarding the loss function, we considered the same proposed in \cite{lutter2019_delan_control}, namely
\begin{equation*}
    \ell = \ell_{id} + \ell_{pc},
\end{equation*}
where $\ell_{id}$ penalized the inverse dynamics error while $\ell_{pc}$ imposes power conservation.
The set of network hyperparameters considered as well as the optimization parameters are reported in Table \ref{tab:Delan_hyperepar}. 
To obtain more accurate results we trained DeLaN models using a larger number of data, obtained sampling the trajectories in \eqref{eq:trajectories_gen} at a frequency of $\SI{100}{\hertz}$, which resulted in datasets composed by 5000 samples. 

\begin{table}[]
    \centering
    \begin{tabular}{|c|c|}
        \hline
          Network parameter & Value \\
         \hline
         \textit{layer width} & 64 \\
         \textit{depth} & 3 \\
         \textit{activation} & \textit{Softplus} \\
         \hline
         Optimization parameter & Value\\
         \hline
         \textit{weight initialization} & \textit{Xavier Normal} \cite{glorot2010understanding}\\
         \textit{minibatch size}& 100\\
         \textit{learning rate} & $5e-4$ \\
         \textit{weight decay} & $1e-5$ \\
         \textit{num epochs} & $2e5$\\
         \hline
    \end{tabular}
    \caption{DeLaN network architecture and Optimization parameters.}
    \label{tab:Delan_hyperepar}
\end{table}

The MC tests described above are performed on different robot configurations, with a number of DOF increasing from \num{3} to \num{7}, to evaluate the performance degradation at the increase of the system dimensionality. The accuracy in predicting the torques is evaluated in terms of normalized Mean Squared Error (nMSE), which provides a measure of the error as a percentage of the signal magnitude. The distribution of the nMSE over the 50 test trajectories for the 6 DOF configuration is shown in Fig.~\ref{fig:gen}. Fig.~\ref{fig:cum_nmse}, instead, summarizes the results obtained on all the considered robot configurations, from 3 to 7 DOF. Results are reported in terms of nMSE percentage averaged over all the joints. 

The proposed \kernelInitials{} model significantly outperforms both the GP-based estimators and the DeLan network. Fig.~\ref{fig:gen} shows that the \kernelInitials{} model provides more accurate results than all the other estimators on all the joints. The improvement is particularly evident on the last two joints which, generally, are more complex to estimate. The \kernelInitials{} is the only model that maintains a nMSE under 10\%. We attribute this performance improvement to the multi-output formulation, combined with the definition of specific basis functions. In this way, the model extrapolates information on the last joints also from the other torque components, thus improving generalization.
The performance of the Delan network decreases faster than the GP-estimators when the number of considered DOF increases, as shown in Table \ref{fig:cum_nmse}. Moreover, from Fig. \ref{fig:gen} we can see that Delan's accuracy is comparable to the one of the GIP and LSE models on the first four joints, with nMSE scores lower than 10\%, but it performs poorly on the last joints where the torque has smaller values.

\begin{figure}[t]
    \centering
    \includegraphics[width=\columnwidth]{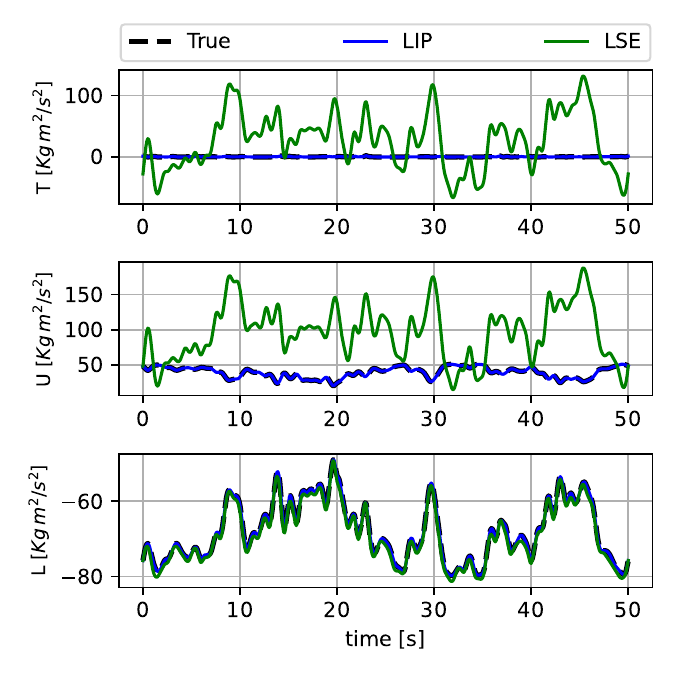}
    \caption{Kinetic energy (T), potential energy (U), and Lagrangian (L) estimated with the LIP and LSE models on one of the trajectories of the MC experiment described in Section \ref{sec:exp_sim}. While the LSE model correctly reconstructs the Lagrangian, it fails to estimate the potential and kinetic energy. In contrast, the LIP model correctly reconstructs all the energy components.
    }
    \label{fig:LSE_energy}
\end{figure}

Then, we tested also the ability of the \kernelInitials{} model to estimate the kinetic and potential energy. The energies are estimated on the same trajectories of the previous MC test by applying the procedure described in section \ref{sec:energy_est}. The distribution of the nMSE error is reported in Fig.~\ref{fig:gen_energy} for the 6 DOF configuration. 
Note that the SE and GIP estimators do not model the system energies and thus do not allow to estimate them. Regarding the LSE estimator, we recall that it models the entire system Lagrangian. In principle, one could derive an estimate of the kinetic and potential energy from the estimate Lagrangian. Let $\hat{\mathcal{L}}(\mathbf{q},\mathbf{\dot{q}})$ the posterior estimate of the Lagrangian, derived following the same procedure described in Section~\ref{sec:energy_est} for the potential and kinetic energies. We can estimate the correspondent potential energy as $\hat{\mathcal{L}}(\mathbf{q},\mathbf{\dot{q}}=0)$, namely, setting velocities to zero. Instead, the kinetic energy estimate can be derived as $\hat{\mathcal{L}}(\mathbf{q},\mathbf{\dot{q}}) + \hat{\mathcal{L}}(\mathbf{q},\mathbf{\dot{q}}=0)$. However, we verified that this procedure returns inaccurate results. As an example, we report in Figure \ref{fig:LSE_energy} the energies estimated with the LSE model on one of the trajectories of the MC experiment for the 6DOF configuration. While the estimation of the overall Lagrangian is quite accurate, the decomposition in kinetic and potential energy is very poor. For this reason, in Fig.~\ref{fig:gen_energy} we compared the performance of the \kernelInitials{} model only with those of the DeLan network.

Results show that the \kernelInitials{} estimator is able to approximate the system energy with high precision. In particular, it reaches a nMSE score close to 1\% on the kinetic energy, while it approximates the potential energy with a nMSE lower than 0.01\%. 
This further confirm the fact that the potential energy, being only function of the joint positions $\q$, is easier to learn, as we observed at the end Section~\ref{subsec:prior_functions}.
The DeLan network provides less accurate performances, which is in accordance with the results obtained in the previous experiments.

\subsubsection{Data efficiency} \label{sec:data_eff}
In the second set of experiments, we tested the data efficiency of the estimators, that is, the estimation accuracy as a function of the number of data points available at training time. 
We collected two datasets, one for training and one for test, using the same type of trajectories as in \eqref{eq:trajectories_gen}. The models are trained increasing the amount of training data from \num{50} to \num{500} samples. Then, their performance  are evaluated on the test dataset.
We carried out this experiment on the 7 DOF configuration of the Panda robot and we compared the \kernelInitials{} model only with the GP-based estimators. The DeLan network has not been considered due to its low accuracy on the 7 DOF configuration.

\begin{figure}[t]
    \centering
    \includegraphics[width=\columnwidth]{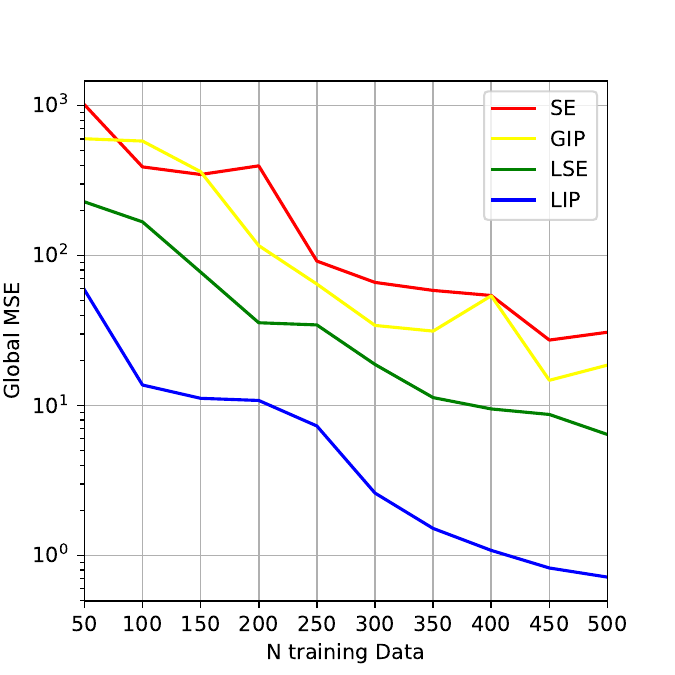}
    \caption{Results of the data efficiency test described in section \ref{sec:data_eff}. The plot shows the evolution of the Global MSE on the test set, as a function of the number of training samples. 
    }
    \label{fig:data_eff}
\end{figure}

Figure \ref{fig:data_eff} reports the evolution of the Global Mean Square Error (Global MSE), namely the sum of the MSE for all the joints, as a function of the number of training samples.
As in the previous experiment, the \kernelInitials{} model outperforms the other GP-based estimators, proving the benefits of the proposed solution also in terms of data efficiency.

\subsection{Real experiments}
\label{sec:real_exp}
The proposed approach is now tested on two real robots: a Franka Emika Panda cobot and an industrial Mitsubishi MELFA RV-4FL. On these robots, we compared the prediction performances of the GP-based estimator considered in section \ref{sec:exp_sim}. 
We modified the GIP, the LSE, and the \kernelInitials{} estimators to account for friction and other unknown effects affecting real systems, i.e., $\bm{\varepsilon} \neq 0$ in \eqref{eq:dyn_eq}. 
We modeled the components of $\bm{\varepsilon}$ as independent zero-mean GPs with covariance function
\begin{equation}
    k^{\varepsilon_i} (\x, \x^\prime) = \phi^i(\x) \Gamma_{\varepsilon_i} \phi^i(\x^\prime)^T + k_{SE}^i(\x, \x^\prime),
\label{eq:kernel_epsilon}
\end{equation}
where the term $\phi^i(\x) \Gamma_{\varepsilon_i} \phi^i(\x^\prime)^T$ is a linear kernel in $\phi^i(\x) = [\dot{q}^i, \, \text{sign}(\dot{q}^i)]^T$, which are the basic features used to describe friction, while $\Gamma_{\varepsilon_i}$ is a diagonal matrix collecting the kernel hyperparameters. The term $k^i_{SE}(\x, \x^\prime)$, instead, is a SE kernel as defined in \eqref{eq:K-SE} and accounts for the remaining unmodeled dynamics. 
For the GIP model, which is based on the single output approach and models the inverse dynamics components independently, we modify the model of the $i$-th  component by adding $k^{\varepsilon_i}(\x, \x^\prime)$ to its kernel.
Concerning the \kernelInitials{} model, the kernel for the inverse dynamics is obtained as 
\begin{equation}
k^{\tau}(\x, \x^\prime) +  \text{diag}(k^{\varepsilon_1}(\x, \x^\prime), \dots, k^{\varepsilon_N}(\x, \x^\prime) ),
\label{eq:kernel_with_fric}
\end{equation}
with $k^{\tau}(\x, \x^\prime)$ being the kernel matrix obtained in \eqref{eq:lagrangian-kernel}. Finally, for the LSE model, we experimentally verified that the addition of $k^i_{SE}$ in \eqref{eq:kernel_epsilon} leads to less accurate results. For this reason, in our implementation, we added to the LSE kernel only the kernel linear in the friction feature, as proposed also in the original paper \cite{cheng-vector-valued-RKHS-4invDyn}.
The resulting kernel is obtain as in\eqref{eq:kernel_with_fric}, with $k^{\varepsilon_i}$ defined as in \eqref{eq:kernel_epsilon}
but without $k^i_{SE}$.

For the experiment performed in this Section, we did not consider the DeLan network due to the poor performance reached in simulation.

\subsubsection{Franka Emika Panda}\label{sec:real_panda}
On the real Panda robot, we collected joint positions, velocities, and torques through the ROS interface provided by the robot manufacturer. To mitigate the effect of measurement noise, we filtered the collected positions, velocities, and torques with a low pass filter with a cut-off frequency of \SI{4}{\hertz}. We obtained joint accelerations by applying acausal differentiation to joint velocities. 

We collected \num{10} training and \num{16} test datasets, following different sum of sinusoids reference trajectories as in \eqref{eq:trajectories_gen}. The test datasets have a wider range of frequencies, $N_s=100$, than the training trajectories, $N_s=50$, to analyze the generalization properties w.r.t. variations of the frequencies used to excite the system. The GP-based estimators are learned in each training trajectory and tested on each and every of the $16$ test trajectories. 

\begin{figure}[ht!]
    \centering
    \includegraphics[width=\columnwidth]{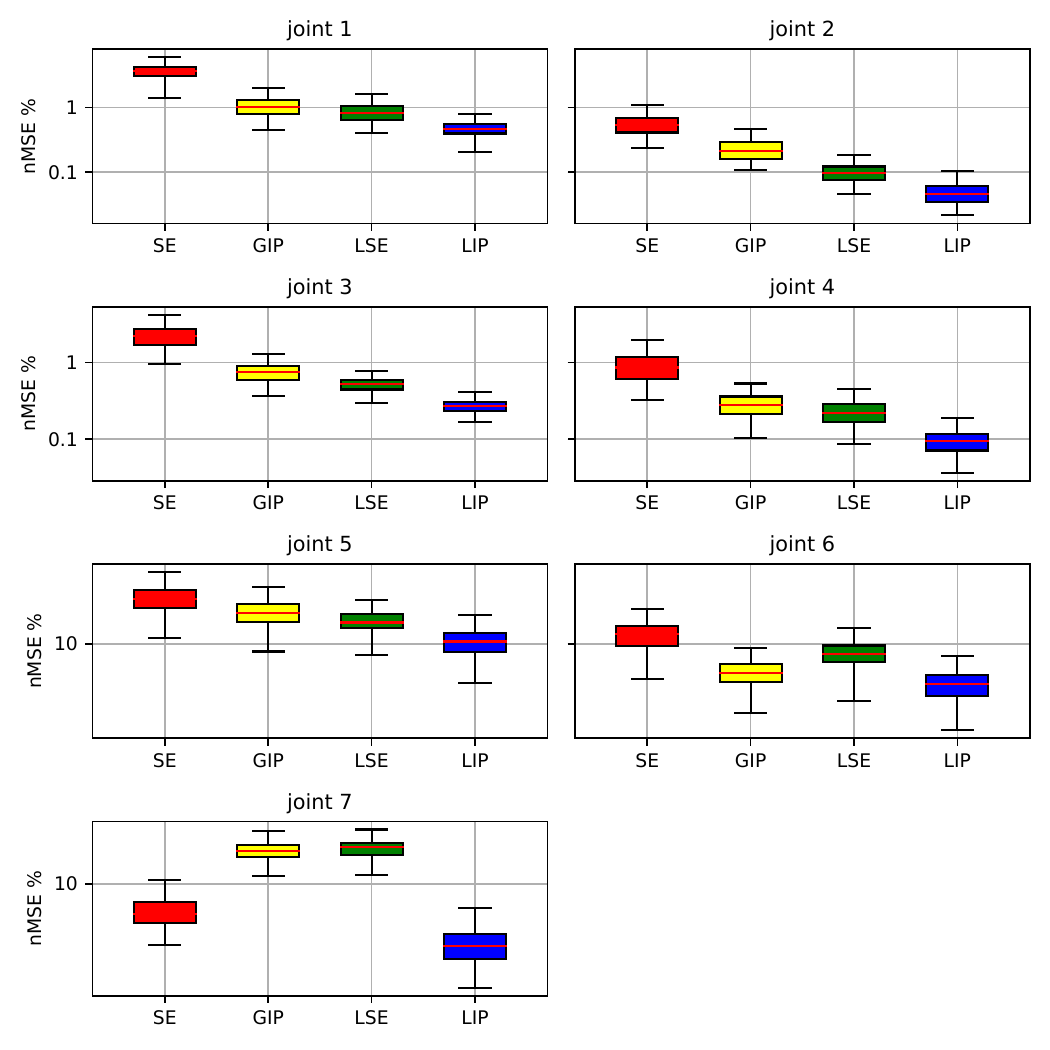}
    \caption[with table]{Box plot of the torque nMSE obtained with the experiment on the PANDA robot described in section \ref{sec:real_panda}. The table below reports the mean nMSE percentage for each joint and each estimator.\\
    
        \centering
        \vspace{5pt}
        \begin{footnotesize}
            \begin{tabular}{ l c c c c c c c}
                \hline \\[-1.5ex]
                
                    & J1 & J2 & J3 & J4 & J5 & J6 & J7\\ [0.5ex]
        
                \hline \\ [-1.5ex]
                SE     & 3.87 & 0.58 & 2.49 & 1.02 & 29.72 & 12.81 & 7.39
                 \\ [0.5ex]
                GIP    & 1.13 & 0.24 & 0.78 & 0.29 & 21.93 & 5.34 & 15.29
                 \\ [0.5ex]
                LSE    &  0.89 & 0.10 & 0.53 & 0.24 & 18.38 & 8.05 & 15.77
                \\ [0.5ex]
                LIP    & \textbf{0.49} & \textbf{0.05} & \textbf{0.28} & \textbf{0.10} & \textbf{11.08} & \textbf{4.10} & \textbf{5.03}
                 \\ [0.5ex]
                \hline
            \end{tabular}
        \end{footnotesize}
        \vspace{5pt}}
    \label{fig:real_panda_bars}
\end{figure}

Figure \ref{fig:real_panda_bars} plots the nMSEs distributions, for each joint. These results confirm the behavior observed in simulation: the \kernelInitials{} model  predicts the joint torques better than other GP estimators on each and every joint, which further proves the advantages of our approach. 

\begin{figure}[t]
    \centering
    \includegraphics[width=\columnwidth]{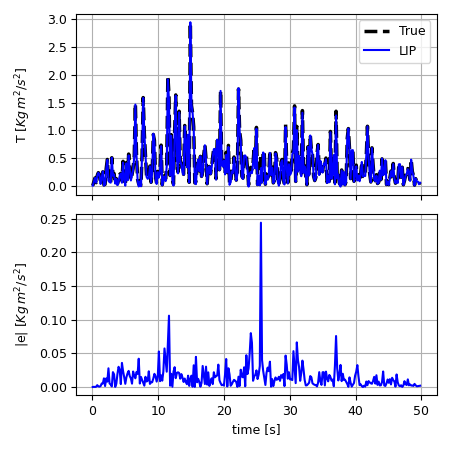}
    \caption{Kinetic energy (upper) and corresponding absolute error (lower) estimated with the \kernelInitials{} model within the experiment described in section \ref{sec:real_panda} on the real Panda robot. The corresponding nMSE is $0.40 \%$.}
    \label{fig:real_panda_energy}
\end{figure}

In Fig.~\ref{fig:real_panda_energy} we report the estimates of the kinetic energy and the corresponding absolute error, obtained in one of the test trajectories. The ground truth has been computed as $T(\q, \dq) = \dq^T B(\q) \dq$, with $B(\q)$ being the inertia matrix at joint configuration $\q$, which is provided by the robot interface. The \kernelInitials{} model accurately predicts the kinetic energy of the system, with an nMSE lower than $1 \%$.

\subsubsection{Mitsubishi MELFA} \label{sec:real_melfa}
The MELFA RV-4FL is a 6 DOF industrial manipulator composed of revolute joints. On this setup, we compared the prediction performance of the \kernelInitials{} model with both black-box and model-based estimators. In particular, we considered the same GP-based estimators as in the previous sections, namely the SE, GIP and LSE models. Concerning the model-based estimators, we implemented two solutions. The first is an estimator obtained with classic Fisherian identification (ID) based on \eqref{eq:linear_model} while the second is a semiparametric kernel-based estimator (SP) with the kernel in \eqref{eq:K-SP}. For both the model-based approaches we computed the matrix $\Phi$ in \eqref{eq:linear_model} using the nominal kinematics and we considered friction contributions. 

\begin{figure}[t]
    \centering
    \includegraphics[width=\columnwidth]{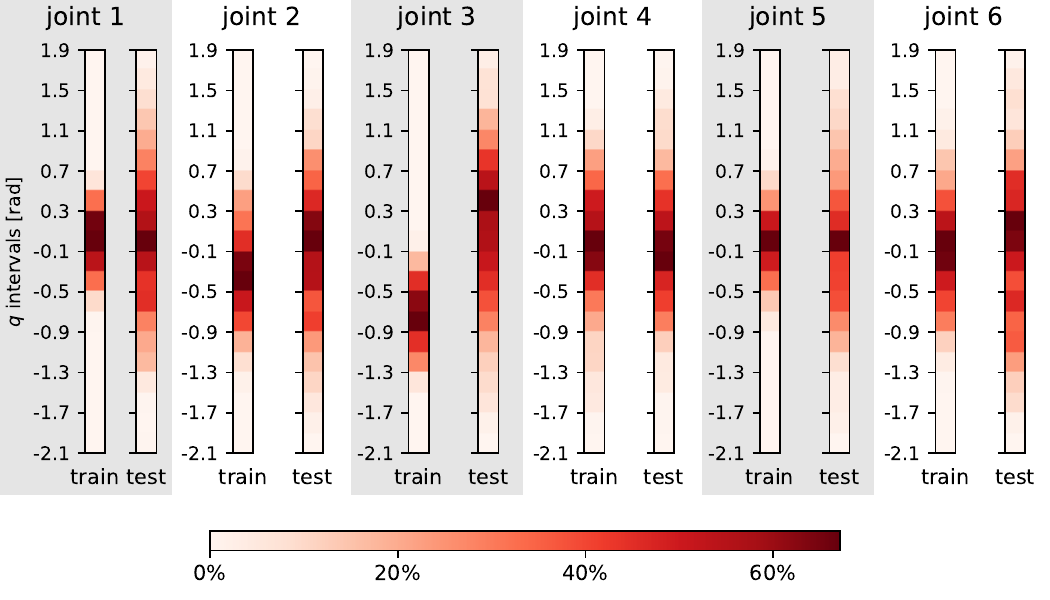}
    \caption{Distribution of the joint positions for the train and test trajectories of the experiment involving the MELFA robot, described in Section \ref{sec:real_melfa}}
    \label{fig:input_dist_melfa}
\end{figure}

We performed the same type of experiment carried out with the Panda robot. In this case, we collected \num{11} training and \num{11} test trajectories. Moreover, both the train and test trajectories are obtained as the sum of $N_s=100$ random sinusoids. To stress generalization we enlarged the position range of the test trajectory. Figure \ref{fig:input_dist_melfa} shows the position distribution of the train and test datasets, from which it can be noticed that the test datasets explore a wider portion of the robot operative range.

\begin{figure}[!ht]
    \centering
    \includegraphics[width=\columnwidth]{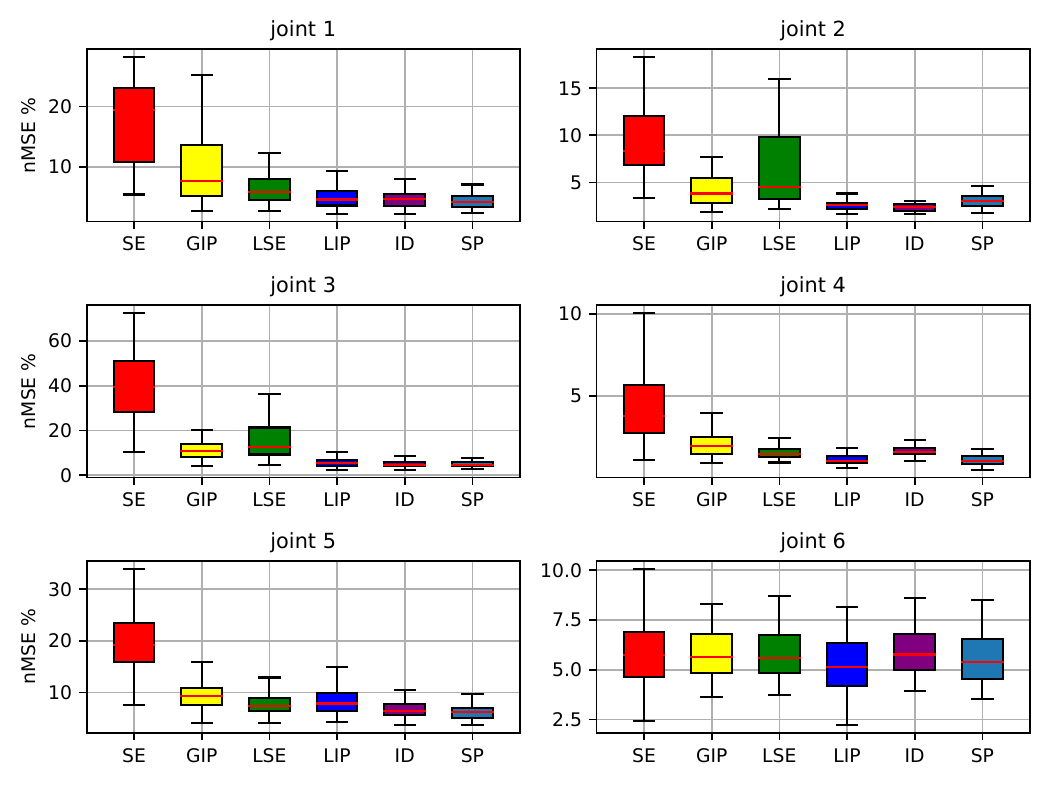}
    \caption[with table]{Box plot of the torque nMSE obtained with the experiment on the MELFA robot described in section \ref{sec:real_melfa}. The table below reports the mean nMSE percentage for each joint and each estimator.\\
    
        \centering
        \vspace{5pt}
        \begin{footnotesize}
            \begin{tabular}{ l c c c c c c}
                \hline \\[-1.5ex]
        
                    & J1 & J2 & J3 & J4 & J5 & J6 \\ [0.5ex]
        
                \hline \\ [-1.5ex]
                SE     & 17.34 &  9.35 & 42.76 & 4.30 & 19.54 & 5.73
                 \\ [0.5ex]
                GIP    & 12.46 & 5.64 & 11.35 & 2.14 & 9.58 & 5.81
                 \\ [0.5ex]
                LSE    &  6.49 & 12.14 & 15.25 & 1.54 & 7.79 &  5.78
                 \\ [0.5ex]
                LIP    & 4.98 & 2.54 & 5.59 & \textbf{1.07} & 8.29 & \textbf{4.74}
                 \\ [0.5ex]
                ID     & 4.86 & \textbf{2.37} & 5.04 & 1.64 & 6.76 & 5.82 
                \\ [0.5ex]
                SP     & \textbf{4.57} & 3.12 & \textbf{5.02} & 1.10 & \textbf{6.44} & 5.49
                \\ [0.5ex]
                \hline
            \end{tabular}
        \end{footnotesize}
        \vspace{5pt}}
    \label{fig:melfa}
\end{figure}

Results in terms of nMSE are reported in Fig.~\ref{fig:melfa}. The \kernelInitials{} model outperforms the other data driven estimators also in this setup. When compared to model-based approaches, the \kernelInitials{} model
reaches results comparable to both the ID and SP models.
Notice that the \kernelInitials{} estimator do not require physical basis functions or parameters and the fact that it performs as well as model-based approaches is a non-trivial achievement. The energy estimation can not be validated in this robot since the ground truth is not available. However, to allow a qualitative evaluation of the energy estimates, in the supplementary material we included a video that shows, simultaneously, the robot following one of the test trajectories and the evolution of the potential and kinetic energy estimates.

\begin{table*}[ht!]
\caption{Joint position nMSE percentage and mean square of the feedback term obtained with the experiments described in section \ref{sec:control_exp} with reference trajectories of different amplitudes.}
\label{tab:control_exp}
\begin{subtable}{1\textwidth}
\caption{Amplitude A = 0.5}
\label{tab:control_0.5}
\centering
\begin{tabular}{ |l| c c c c c c c| c c c c c c c|}
 \hline
 \multicolumn{1}{|c|}{\multirow{2}{*}{model}}&
 \multicolumn{7}{|c|}{Joint position nMSE \%} &
 \multicolumn{7}{|c|}{Feedback term mean square [$N^2m^2$]} \\
 
 \cline{2-15}
 \multicolumn{1}{|c|}{}
 & J1 & J2 & J3 & J4 & J5 & J6 & J7& J1 & J2 & J3 & J4 & J5 & J6 & J7\\ 
 \hline 
 LIP    & \footnotesize{5.1e-4}
        & \footnotesize{1.0e-3} 
        & \footnotesize{3.3e-3} 
        & \footnotesize{3.0e-3} 
        & \footnotesize{9.4e-3} 
        & \footnotesize{5.9e-3} 
        & \footnotesize{6.0e-3}
        & \footnotesize{1.9e-1}
        & \footnotesize{3.5e-1} 
        & \footnotesize{2.1e-1} 
        & \footnotesize{1.8e-1} 
        & \footnotesize{0.5e-1} 
        & \footnotesize{0.2e-1} 
        & \footnotesize{0.1e-1}\\        
 nom    & \footnotesize{6.5e-3} 
        & \footnotesize{7.9e-3} 
        & \footnotesize{3.0e-2} 
        & \footnotesize{7.2e-2} 
        & \footnotesize{2.3e-1} 
        & \footnotesize{1.6e-1} 
        & \footnotesize{9.0e-2}
        & \footnotesize{1.5e0} 
        & \footnotesize{1.7e0} 
        & \footnotesize{1.6e0} 
        & \footnotesize{2.3e0}
        & \footnotesize{1.0e0}
        & \footnotesize{3.3e-1}
        & \footnotesize{1.9e-1}\\
 \hline
\end{tabular}
\vspace{10pt}
\end{subtable}

\begin{subtable}{1\textwidth}
\caption{Amplitude A = 0.6}
\label{tab:control_0.6}
\centering
\begin{tabular}{ |l| c c c c c c c| c c c c c c c|}
 \hline 
 \multicolumn{1}{|c|}{\multirow{2}{*}{model}}&
 \multicolumn{7}{|c|}{Joint position nMSE \%} &
 \multicolumn{7}{|c|}{Feedback Mean Square [$N^2m^2$]} \\
 \cline{2-15}
 \multicolumn{1}{|c|}{}
 & J1 & J2 & J3 & J4 & J5 & J6 & J7& J1 & J2 & J3 & J4 & J5 & J6 & J7\\ 
 \hline 
 LIP    & \footnotesize{7.1e-4}
        & \footnotesize{1.6e-3}
        & \footnotesize{5.6e-3}
        & \footnotesize{6.3e-3}
        & \footnotesize{1.0e-2}
        & \footnotesize{7.4e-3}
        & \footnotesize{9.7e-3}
        & \footnotesize{3.2e-1}
        & \footnotesize{5.0e-1}
        & \footnotesize{3.6e-1}
        & \footnotesize{4.0e-1}
        & \footnotesize{0.6e-1}
        & \footnotesize{0.2e-1}
        & \footnotesize{0.2e-1}\\
 nom    & \footnotesize{6.8e-3}
        & \footnotesize{7.9e-3}
        & \footnotesize{3.1e-2}
        & \footnotesize{6.6e-2}
        & \footnotesize{2.4e-1}
        & \footnotesize{1.6e-1}
        & \footnotesize{8.2e-2}
        & \footnotesize{1.7e0}
        & \footnotesize{1.8e0}
        & \footnotesize{1.8e0}
        & \footnotesize{2.3e0}
        & \footnotesize{1.1e0}
        & \footnotesize{3.6e-1}
        & \footnotesize{1.9e-1}\\
 \hline
\end{tabular}
\end{subtable}
\end{table*}

\begin{figure*}[h!]
    \centering
    \includegraphics[width=\textwidth]{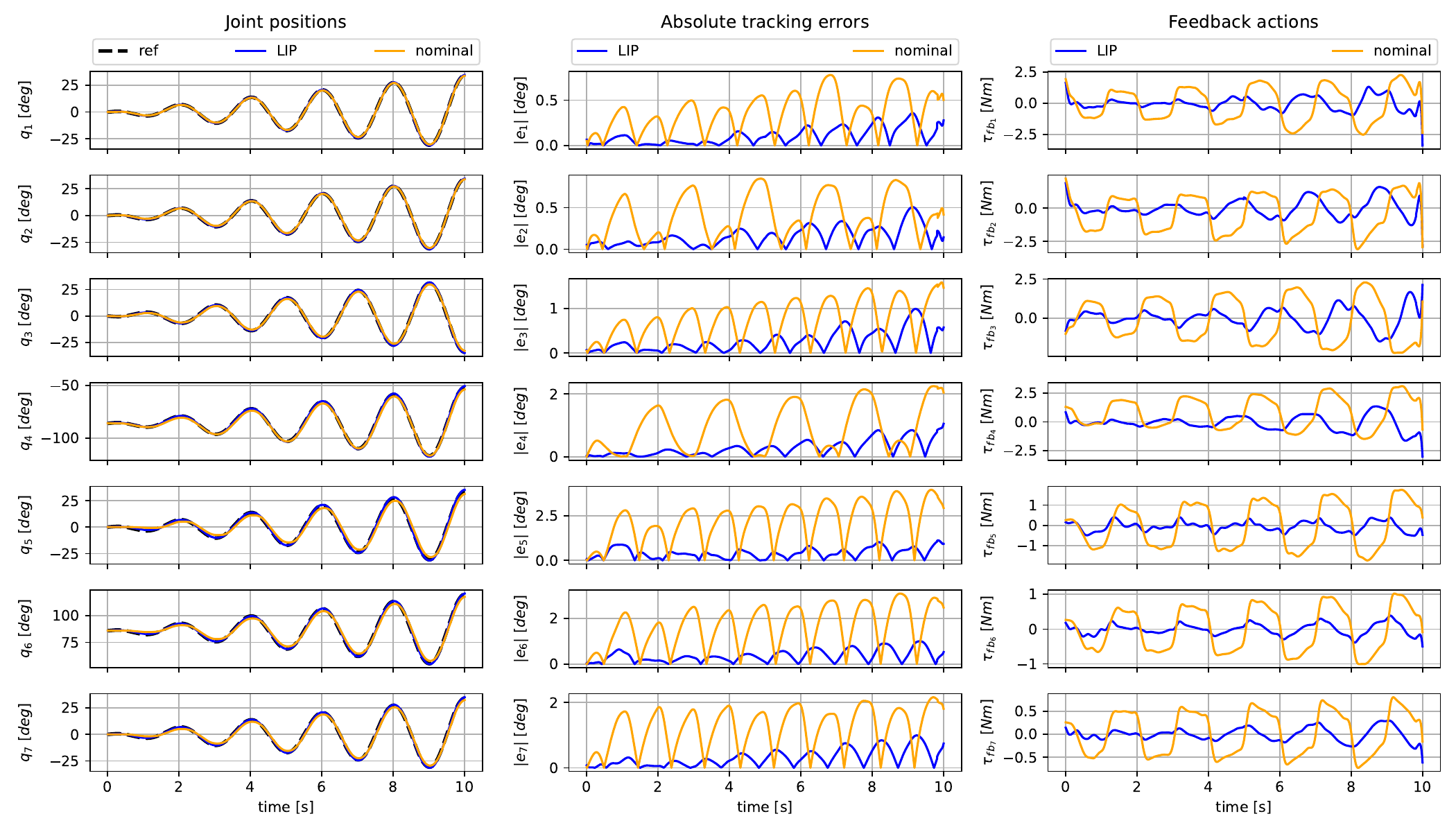}
    \caption{Trajectories obtained with the tracking experiment described in section \ref{sec:control_exp}. The first column reports joint positions, the second column reports the absolute value of the tracking error, while the third column reports the PD action.}
    \label{fig:control_exp}
\end{figure*}

\subsection{Trajectory tracking experiment}
\label{sec:control_exp}
In this section we exploit the LIP model to implement a Computed Torque (CT) control strategy on the real Panda robot.  
Let $\q_d$, $\dq_d$, $\ddq_d$ be the reference trajectory. The CT control law is given by 
\begin{equation}
    \label{eq:CT_control}
    \taubf_{ct}= \taubf_{ff} + \taubf_{fb}, 
\end{equation}
where the feedforward $\taubf_{ff}$ exploits the inverse dynamics to predict the joint torques required to follow the reference trajectory $ (\q_d, \dq_d, \ddq_d)$, while the feedback $\taubf_{fb}$ ensures robustness to modeling error and external disturbances.  We consider a linear feedback control, namely $\taubf_{fb} = K_p \mathbf{e} + K_d \dot{\mathbf{e}}$, where $\mathbf{e} = \q_d - \q$ denotes the tracking error, while the gains $K_p$ and $K_d$ are positive definite diagonal matrices.
For each joint, the reference trajectory is  a sinusoid with a linear envelop, defined as
\begin{equation}
    \label{eq:ref_ctrl}
    q^i_d(t) = \frac{A}{T} \, t \, \text{cos}(\pi\,t) 
\end{equation}
where $A$ represents the maximum amplitude and $T$ the control horizon. For all the performed experiments, we considered $T=\SI{10}{\second}$. All the controllers have been implemented exploiting the torque control interface provided by the robot and run at a frequency of $\SI{1}{\kilo\hertz}$.

The LIP model have been used to compute the feedforward term, namely $\taubf_{ff}=\hat{f}(\q_d, \dq_d, \ddq_d)$, where $\hat{f}$ is the LIP estimation, obtained through \eqref{eq:GP-posterior}.
To collect the training dataset, we run the CT controller with the feedforward computed using the nominal model provided by the robot manufacturer, on a reference trajectory with $A= \SI{0.5}{\radian}$. The collected trajectory consists of $T = \SI{10}{\second}$ sampled at a frequency of $\SI{1}{\kilo\hertz}$, resulting in \SI{10000} samples. In order to reduce the computational burden, we down-sampled the data with a constant rate of $20$, to obtain a dataset of $500$ samples.

Results are summarized in Table \ref{tab:control_exp} and in Fig. \ref{fig:control_exp}. First, in Table~\ref{tab:control_0.5} we report the performance on the trajectory \eqref{eq:ref_ctrl} with $A= \SI{0.5}{\radian}$, in terms of tracking error nMSE and feedback mean square. On this trajectory, the LIP estimator provides a more accurate feedforward than the nominal model, which leads to a sensible reduction of the tracking error. Moreover, the better feedforward induced by the LIP kernel gives benefits also in terms of lower effort required to the feedback action.

To stress out-of-sample generalization, we also tested the tracking accuracy on a trajectory with amplitude $A= \SI{0.6}{\radian}$, which explores a broader position range with higher accelerations and velocities.
In~Table \ref{tab:control_0.6} we summarize the nMSE of the tracking error and the mean square of the feedback action, while in Fig.~\ref{fig:control_exp} we depict the the evolution of the joint angles, absolute tracking errors and feedback action obtained with the LIP and nominal models, respectively. 
Comparing the results, we can observe that the LIP model maintains a similar level of performance, both in terms of tracking accuracy and effort required to the feedback action. This further confirms the good generalization properties of the proposed approach already highlighted in section \ref{sec:gen}.

\subsection{Discussion}
In both simulated and real-world setups, the LIP model consistently outperformed other state-of-the-art black-box estimators. When contrasted with conventional model-based identification approaches, as demonstrated in the MELFA robot experiments detailed in Section~\ref{sec:real_melfa}, the LIP model exhibited comparable performance, without requiring any prior knowledge of the system. These outcomes further confirm the soundness of the modeling strategies proposed, that we discuss in the following.

First, encoding the Euler Lagrange equations endows the model with structural properties that aligns with physical principles. Consequently, this approach enhances the model's ability to interpret the information contained within the data, yielding benefits in both out-of-sample generalization and data efficiency.

More importantly, a substantial enhancement in performance is given by the specific choice of energy priors. Indeed, from a functional analysis perspective, the choice of the kernel is strictly related to the basis functions of the hypothesis space. The proposed polynomial kernels induce a finite set of polynomial basis functions, specifically tailored to the inverse dynamics problem. 
In contrast, universal kernels, such as the SE, induce a broader hypothesis space with an infinite number of basis functions, thereby increasing the complexity of the learning problem. The benefits of the proposed priors become apparent when comparing the performance of the LIP estimator with that of the LSE model.

Finally, the DeLaN baseline exhibited lower accuracy compared to all GP counterparts in our examined scenarios. This outcome is expected, given that Neural Network models, like DeLaN, generally require more data compared to GPR approaches, and in our experiments, we considered relatively small datasets. Additionally, the inherent complexity of structured networks such as DeLaN poses challenges in training, as noted in previous research \cite{cranmer2019lagrangian}. Despite efforts to mitigate this issue by training DeLaN models with larger datasets, the observed performance degradation with an increase in system DOF suggests that data richness remains insufficient. This underscores the advantage of GPR in application with limited data availability, owing to its inherent data efficiency.

A potential drawback of GP models is the high computational burden required for model training. The time required to train a GP model scales cubically with the number of training samples. 
Online evaluation, instead, scales linearly with both the number of training samples and the number of DOFs. However, an efficient implementation of the kernel computation plays a fundamental role when considering time complexity. Conveniently, most of the operations can be vectorized and leverage hardware solutions such as GPUs. This aspect has not been fully explored in this work and needs to be addressed when considering time-critical applications, such as online control. Furthermore, it is worth mentioning that several approximation strategies have been proposed to reduce both training and evaluation time. Among these, Variational Inference (VI) \cite{leibfried2020tutorial} has shown interesting results. We leave the investigation of VI applied to the proposed method as a subject of future work.

Another critical aspect when considering black-box estimators is physical consistency. Physical consistency refers to the compliance of learned models with known physical principles, such as symmetries and conservation laws. Since the LIP kernel is derived by imposing the Euler-Lagrange equations, all the properties related to the structure of such equations are inherited by the model. As an example, this means that the symmetry of the inertia matrix and the skew-symmetry of the Coriolis matrix are imposed by design. Nonetheless, other relevant properties, such as the positive definiteness of the inertia matrix, are not imposed and thus are not guaranteed to be respected. In turn, non-positive definite inertial matrices can lead to instability issues when employed for control purposes. To mitigate this problem, constrained optimization techniques to learn either the hyperparameters or the posterior coefficients imposing positivity constraints represent an appealing solution \cite{swiler2020survey}, which we are considering as an extension of this work.

\section{Conclusions}\label{sec:conclusions}

In this work, we presented the \kernelInitials{} model, a GPR estimator based on a novel multimensional kernel, designed to model the kinetic and potential energy of the system. The proposed method has been validated both on simulated and real setups involving a Franka Emika Panda robot and a Mitsubishi MELFA RV-4FL. The collected results showed that it outperforms state-of-the-art black box estimators based on Gaussian Processes and Neural-Networks in terms of data efficiency and generalization. Moreover, results on the MELFA robot demonstrated that our method achieves a prediction accuracy comparable to the one of fine-tuned model-based estimators, despite requiring less prior information. Finally, the experiments in simulation and on the real Panda robot proved the effectiveness of the \kernelInitials{} model also in terms of energy estimation.

\section*{Acknowledgements}
Alberto Dalla Libera was supported by PNRR research activities of the consortium iNEST (Interconnected North-Est Innovation Ecosystem) funded by the European Union Next-GenerationEU (Piano Nazionale di Ripresa e Resilienza (PNRR) – Missione 4 Componente 2, Investimento 1.5 – D.D. 1058  23/06/2022, ECS\_00000043). This manuscript reflects only the Authors’ views and opinions, neither the European Union nor the European Commission can be considered responsible for them.

\appendix
In this appendix, we provide the proofs of Propositions \ref{prop:pot_pol} and \ref{prop:kin_pol}. The proofs follow from similar concepts exploited in \cite{ADL_GIP19}. For convenience, we consider first Proposition \ref{prop:kin_pol}, since most of the elements we introduce in the corresponding proof are useful also to prove Proposition \ref{prop:pot_pol}.
\subsection{Proof of Proposition \ref{prop:kin_pol}}
We start by recalling that the kinetic energy of the $i$-th link can be written as
\begin{equation}\label{eq:kin_i}
    \mathcal{T}_i =  \dq^{i T} B_i(\q^i) \dq^i, 
\end{equation}
with $B_i(\q^i) \in \R ^ {i \times i}$. Then, we characterize the elements of $B_i(\q^i)$ as polynomial functions in $\q_c^i$, $\q_s^i$ and $\q_p^i$.
The matrix $B_i(\q^i)$ is defined as 
\begin{equation}
    \label{eq:def_of_B_i}
    B_i(\q^i) = m_i J_{P_i}^{T} J_{P_i} + J_{O_i}^T R_i^0 I_i^i R_0^i J_{O_i}
\end{equation}
where $m_i$ is the mass of the $i$-th link, while $I_i^i$ is its inertia matrix expressed w.r.t. a reference frame (RF) integral with the $i$-th link itself, denominated hereafter $i$-th RF, see \cite[Ch. 7]{siciliano}. We assume that the reference frames have been assigned based on Denavit-Hartenberg (DH) convention.
$J_{P_i}$ and $J_{O_i}$ are the linear and angular Jacobians of the $i$-th RF, for which it holds that $\dot{\boldsymbol{c}}_i = J_{P_i} \dq^i$ and $\boldsymbol{\omega}_i = J_{O_i} \dq^i$, where $\boldsymbol{c}_i$ is the position of the center of mass of the $i$-th link, while $\boldsymbol{\omega}_i$ is the angular velocity of the $i$-th RF w.r.t. the base RF. $R_i^0$ denotes the rotation matrix of the $i$-th RF w.r.t. the base RF.

For later convenience, we recall some notions regarding kinematics. 
Let $R_i^{i-1}$ and $\boldsymbol{l}_i^{i-1}$ denote the orientation and translation of the $i$-th RF w.r.t. the previous one. 
Based on the Denavit-Hartenberg (DH) convention, it is known that $R_i^{i-1}$ and $\boldsymbol{l}_i^{i-1}$ have the following expressions
\begin{align*}
    \label{eq:rel_rototrasl_i}
    R_i^{i-1} &= R_z(\theta_i)R_x(\alpha_i),\\
    \boldsymbol{l}_i^{i-1} &= [0, 0, d_i]^T + R_z(\theta_i)[a_i, 0, 0]^T, 
\end{align*}
where $R_x$ and $R_z$ represent elementary rotations around the $x$ and $z$ axis, respectively, $a_i$ and $\alpha_i$ are constant kinematics parameters depending on the geometry of Link $i$, while
the values of $d_i$ and $\theta_i$ depends on the type of the $i$-th joint; see \cite[Ch. 2]{siciliano} for a detailed discussion. If the $i$-th joint is revolute, then $d_i$ is constant and $\theta_i = \theta_{0_i} + q_i$, where $q_i$ is the $i$-th generalize coordinate, that is, the $i$-th component of the vector $\q$. In this case, the only quantity depending on $\q$ is the rotation matrix $R_i^{i-1}$, whose elements contain the terms $\cos(q_i)$ and $\sin(q_i)$. Therefore, using the notation introduced in Section \ref{subsec:prior_functions}, the elements of $R_i^{i-1}$ can be written as functions in the space $\mathbb{P}_{(1)}(\cos(q_i)_{(1)}, \sin(q_i)_{(1)})$. Instead, if the $i$-th joint is prismatic, $\theta_i$ is constant and $d_i = d_{0_i} + q_i$. The only terms depending on $\q$ are the elements of $\boldsymbol{l}_i^{i-1}$, which belong to the space $\mathbb{P}_{(1)}(q_{i_{(1)}})$. 

Concerning the angular Jacobian $J_{O_i}$, we have that $\boldsymbol{\omega}_i = \sum_{j=1}^{i} R_{j-1}^0 \boldsymbol{\omega}_j^{j-1}$, with $R_{j}^0= \prod_{b=1}^{j} R_b^{b-1}$. Adopting the DH convention, $\boldsymbol{\omega}_i^{i-1} = \lambda_i[0, 0, \dot{q}_i]^T$, with $\lambda_i = 1$ if joint $i$ is revolute and $\lambda_i=0$ if it is prismatic. The expression of $\boldsymbol{\omega}_i$ can be rewritten as
\begin{equation}
\label{eq:omega_i}
\boldsymbol{\omega}_i = 
\begin{bmatrix}
    R_0^0 
    \begin{bmatrix}
        0 \\ 0 \\ \lambda_1    
    \end{bmatrix}
    , \dots,
    R_{i-1}^0 
    \begin{bmatrix}
        0 \\ 0 \\ \lambda_i    
    \end{bmatrix}
\end{bmatrix}
\dq^i.
\end{equation}
From \eqref{eq:omega_i} we deduce that
\begin{equation}
    J_{O_i} = 
    \begin{bmatrix}
    R_0^0 
    \begin{bmatrix}
        0 \\ 0 \\ \lambda_1    
    \end{bmatrix}
    , \dots,
    R_{i-1}^0 
    \begin{bmatrix}
        0 \\ 0 \\ \lambda_i    
    \end{bmatrix}
\end{bmatrix},
\end{equation}
from which we obtain that the term $R_0^iJ_{O_i}$ has expression
\begin{equation}
    R_0^iJ_{O_i} = 
    \begin{bmatrix}
    R_0^i 
    \begin{bmatrix}
        0 \\ 0 \\ \lambda_1    
    \end{bmatrix}
    , \dots,
    R_{i-1}^i 
    \begin{bmatrix}
        0 \\ 0 \\ \lambda_i    
    \end{bmatrix}
\end{bmatrix}.
\end{equation}
Recalling that $R_j^k = \prod_{b=k}^j R_b^{b-1}$, with $j>k$, we have that the elements of $J_{O_i}^T R_i^0 I_i^i R_0^i J_{O_i}$ belong to $\mathbb{P}_{(2|I_r^i|)}(\q_{c_{(2)}}^i, \q_{s_{(2)}}^i)$, where $|I_r^i|$ is the cardinality of $I_r^i$,  and they are composed of monomials with 
\begin{equation}
\label{eq:deg_constraint_1}
    deg(q_{c}^b) + deg(q_{s}^b) \leq 2.
\end{equation}

Similarly, the elements of $J_{P_i}$ are characterized by analyzing the expression $\boldsymbol{c}_i = \sum_{j=1}^{j=i-1}R_{j-1}^0\boldsymbol{l}_j^{j-1} + R_i^0\boldsymbol{c}_i^i$. Then, the elements of $\boldsymbol{c}_i$ are functions in $\mathbb{P}_{(i)}(\q_{c_{(1)}}^i, \q_{s_{(1)}}^i,\q_{p_{(1)}}^i)$ and for each monomial it holds that
\begin{equation}
\label{eq:deg_constraint}
    deg(q_{c}^b) + deg(q_{s}^b) \leq 1.
\end{equation}
Recalling that $\dot{\boldsymbol{c}}_i = J_{P_i} \dq$ and since the derivative does not increase the degree, we can conclude that the elements of $J_{P_i}$ belong to the same polynomial space of $\boldsymbol{c}_i$. As a consequence, the elements of $J_{P_i}^TJ_{P_i}$ belong to $\mathbb{P}_{(2i)}(\q_{c_{(2)}}^i, \q_{s_{(2)}}^i,\q_{p_{(2)}}^i)$ with
\begin{equation}
\label{eq:deg_constraint_3}
    deg(q_{c}^b) + deg(q_{s}^b) \leq 2.
\end{equation}

From the above characterizations of $J_{P_i}^TJ_{P_i}$ and $J_{O_i}^T R_i^0 I_i^i R_0^i J_{O_i}$ we obtain that the elements of $B_i(\q^i)$ belong to $\mathbb{P}_{(2i)}(\q_{c_{(2)}}^i, \q_{s_{(2)}}^i,\q_{p_{(2)}}^i)$. Therefore, it is trivial to see that $T_i(\q,\dq) = \dq^{iT} B_i(\q^i) \dq^i$ belongs to $\mathbb{P}_{(2i+2)}(\q_{c_{(2)}}^i, \q_{s_{(2)}}^i,\q_{p_{(2)}}^i,\dq^i_{(2)})$ with the constraint expressed in \eqref{eq:deg_constraint_3}, which concludes the proof.

\subsection{Proof of Proposition \ref{prop:pot_pol}}
The potential energy $V(\q)$ is defined as
\begin{equation}
    \mathcal{V}(\q) = \sum_{i=1}^{n}m_i\g_0^T \boldsymbol{c}_i,   
\end{equation}
where $\g_0$ is the vector of the gravitational acceleration. From the expression above it follows that $\mathcal{V}$ belongs to the same space of the elements of $\boldsymbol{c}_n$, namely it is a function in $\mathbb{P}_{(n)}(\q_{c_{(1)}}, \q_{s_{(1)}},\q_{p_{(1)}})$, with 
\begin{equation}
\label{eq:deg_constraint_potential}
    deg(q_{c}^b) + deg(q_{s}^b) \leq 1
\end{equation}
for each monomial of the aforementioned polynomial,
which concludes the proof.

\bibliographystyle{IEEEtran}
\bibliography{references}

\newpage

\begin{IEEEbiography}[{\includegraphics[width=1in,height=1.25in,clip,keepaspectratio]{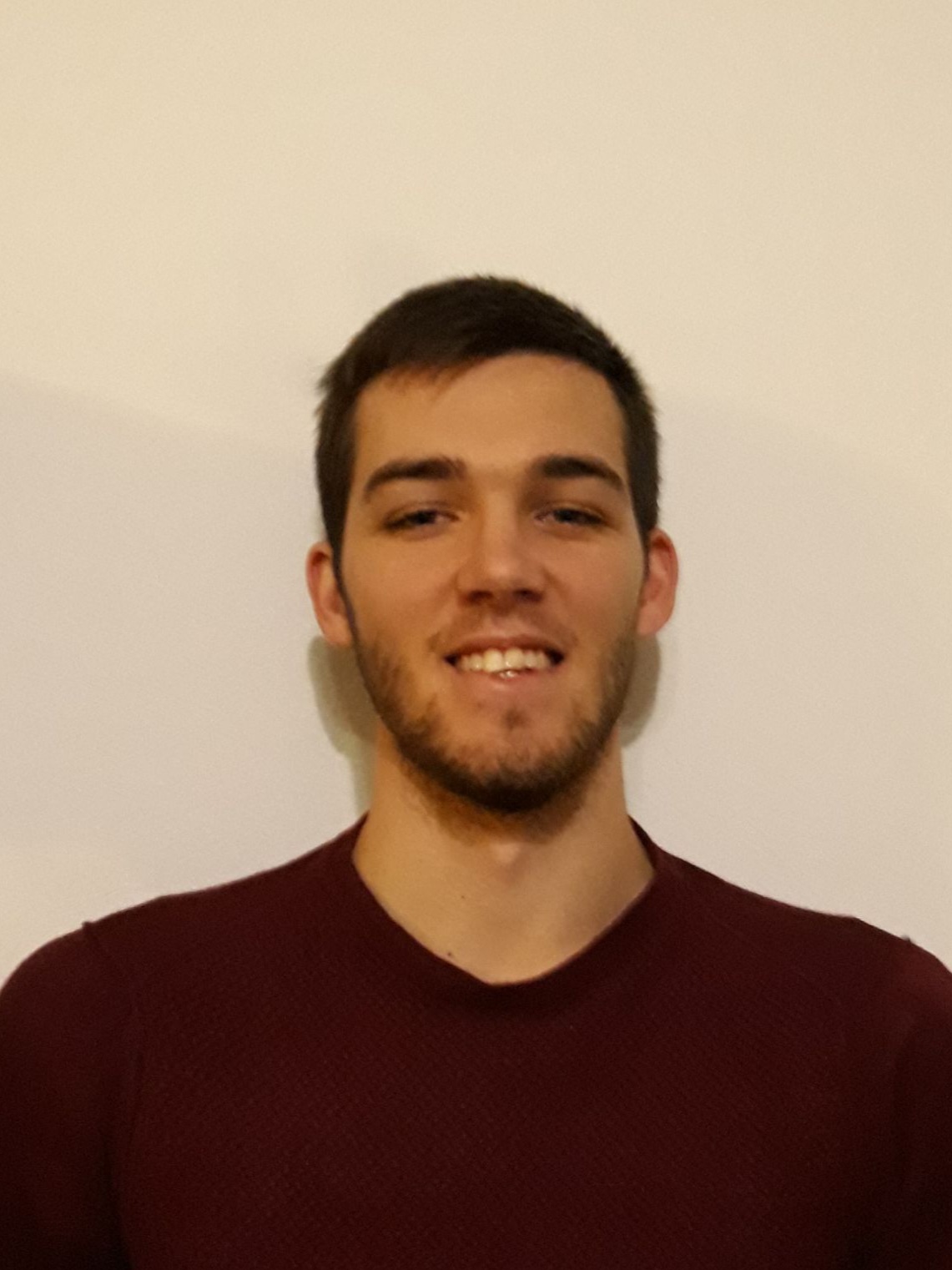}}]{Giulio Giacomuzzo}
Giulio Giacomuzzo received the B.Sc. degree in Information Engineering and the M.Sc. in Control Engineering from the University of Padova, Padua, Italy, in 2017 and 2020, respectively. 

He is currently a Ph.D. student at the Department of Information Engineering, University of Padova. His research interests include Robotics, Machine Learning, Reinformcement Learning and Identification theory. 
\end{IEEEbiography}

\begin{IEEEbiography}[{\includegraphics[width=1in,height=1.25in,clip,keepaspectratio]{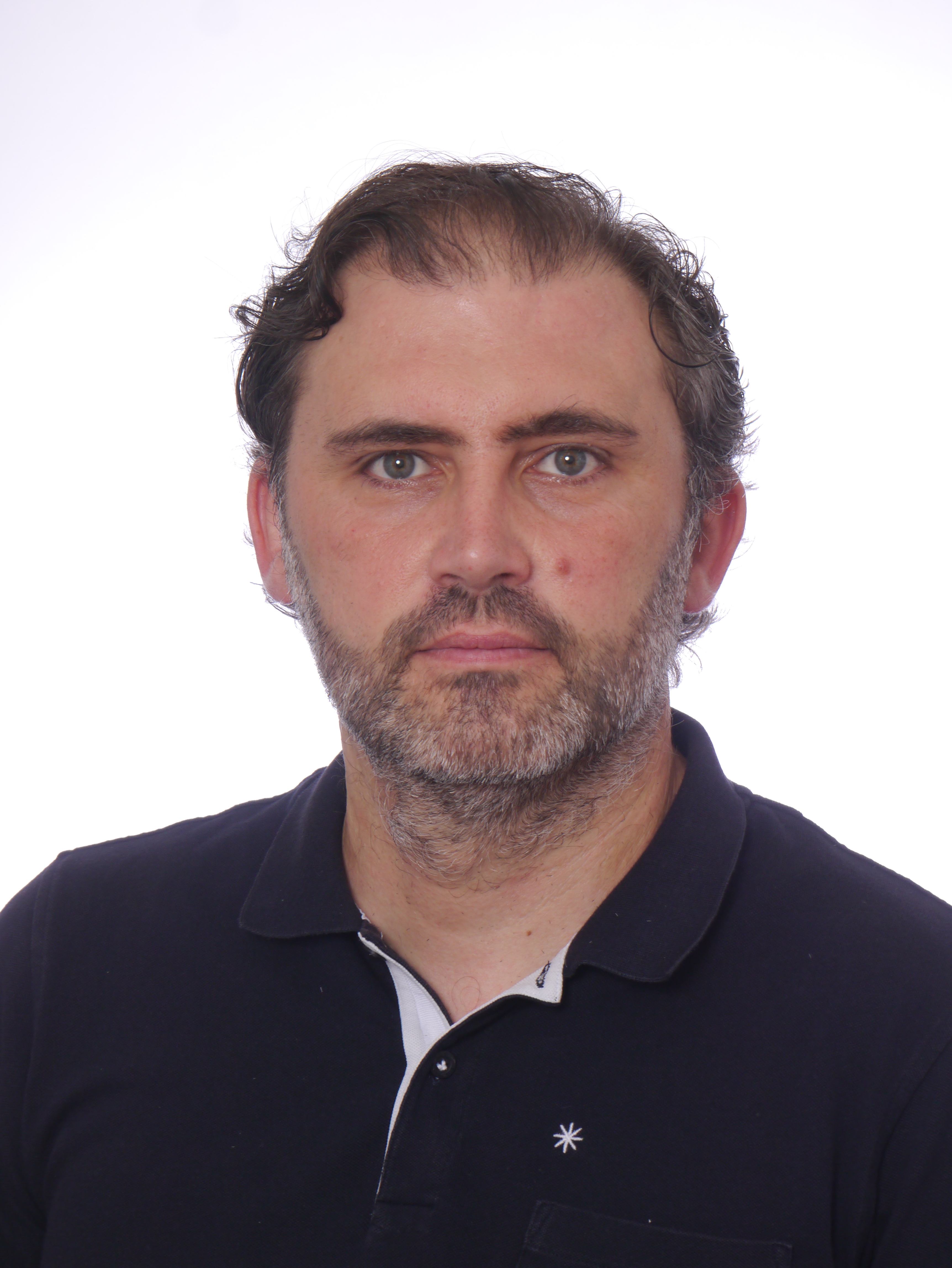}}]{Ruggero Carli}
Ruggero Carli received the Laurea degree in Computer Engineering and the Ph.D. degree in Information Engineering from the University of Padova, Padua, Italy, in 2004 and 2007, respectively. 

From 2008 to 2010, he was a Postdoctoral Fellow with the Department of Mechanical Engineering, University of California at Santa Barbara, Santa Barbara, CA, USA. He is currently an Associate Professor with the Department of Information Engineering, University of Padova. His research interests include distributed algorithms for optimization, estimation and control over networks, nonparametric estimation and learning for robotics.
\end{IEEEbiography}

\begin{IEEEbiography}[{\includegraphics[width=1in,height=1.25in,clip,keepaspectratio]{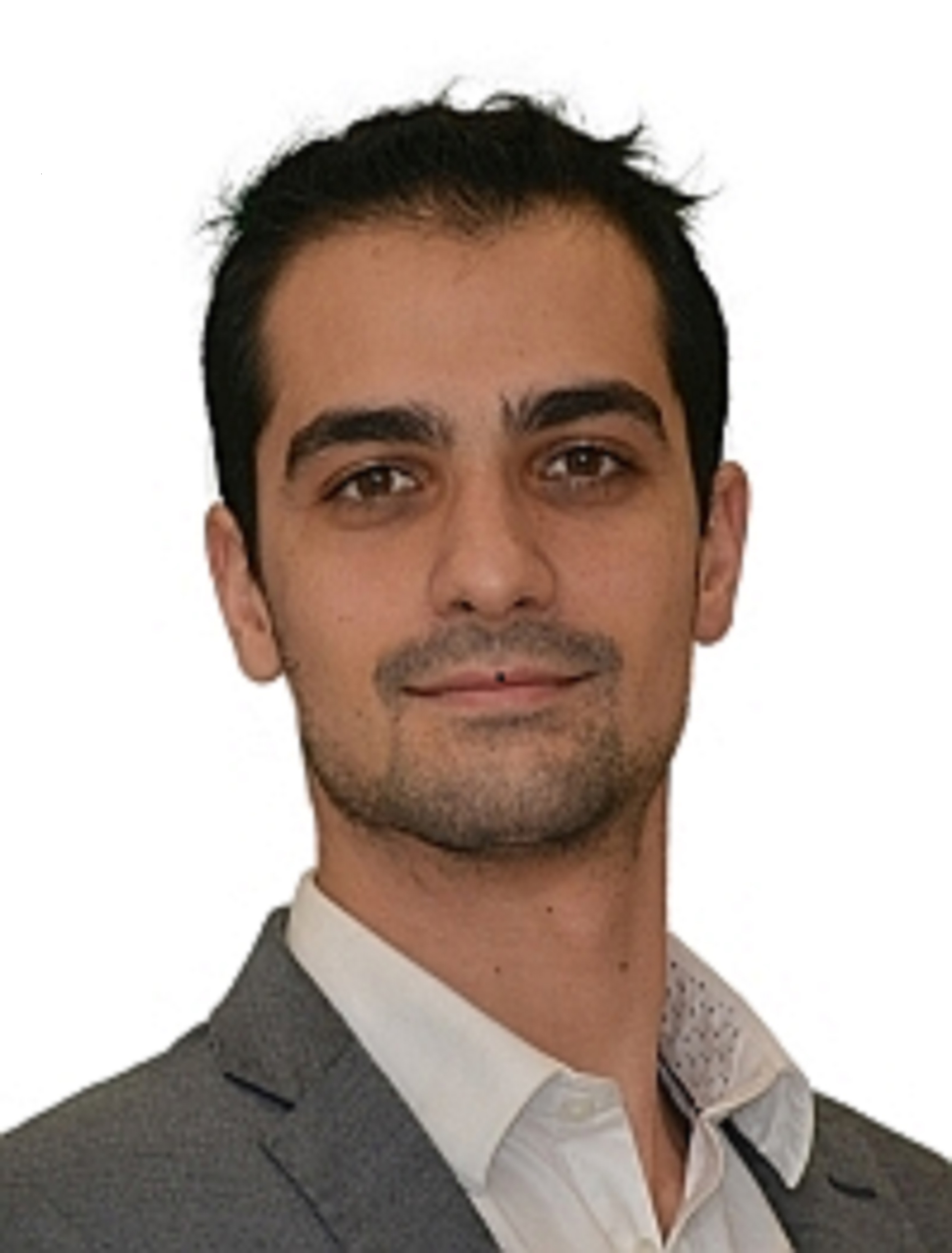}}]{Diego Romeres}
 received the M.Sc. degree (summa cum laude) in control engineering and the Ph.D. degree in information engineering from the University of Padova, Padua, Italy, in 2012 and 2017, respectively.

He is currently a Senior Principal Research Scientist and Team Leader at Mitsubishi Electric Research Laboratories, Cambridge, MA, USA. He held visiting research positions with TU Darmstadt, Germany, and with ETH, Zurich, Switzerland. His research interests include robotics, human-robot interaction, machine learning, bayesian optimization, and system identification theory.
\end{IEEEbiography}

\begin{IEEEbiography}[{\includegraphics[width=1in,height=1.25in,clip,keepaspectratio]{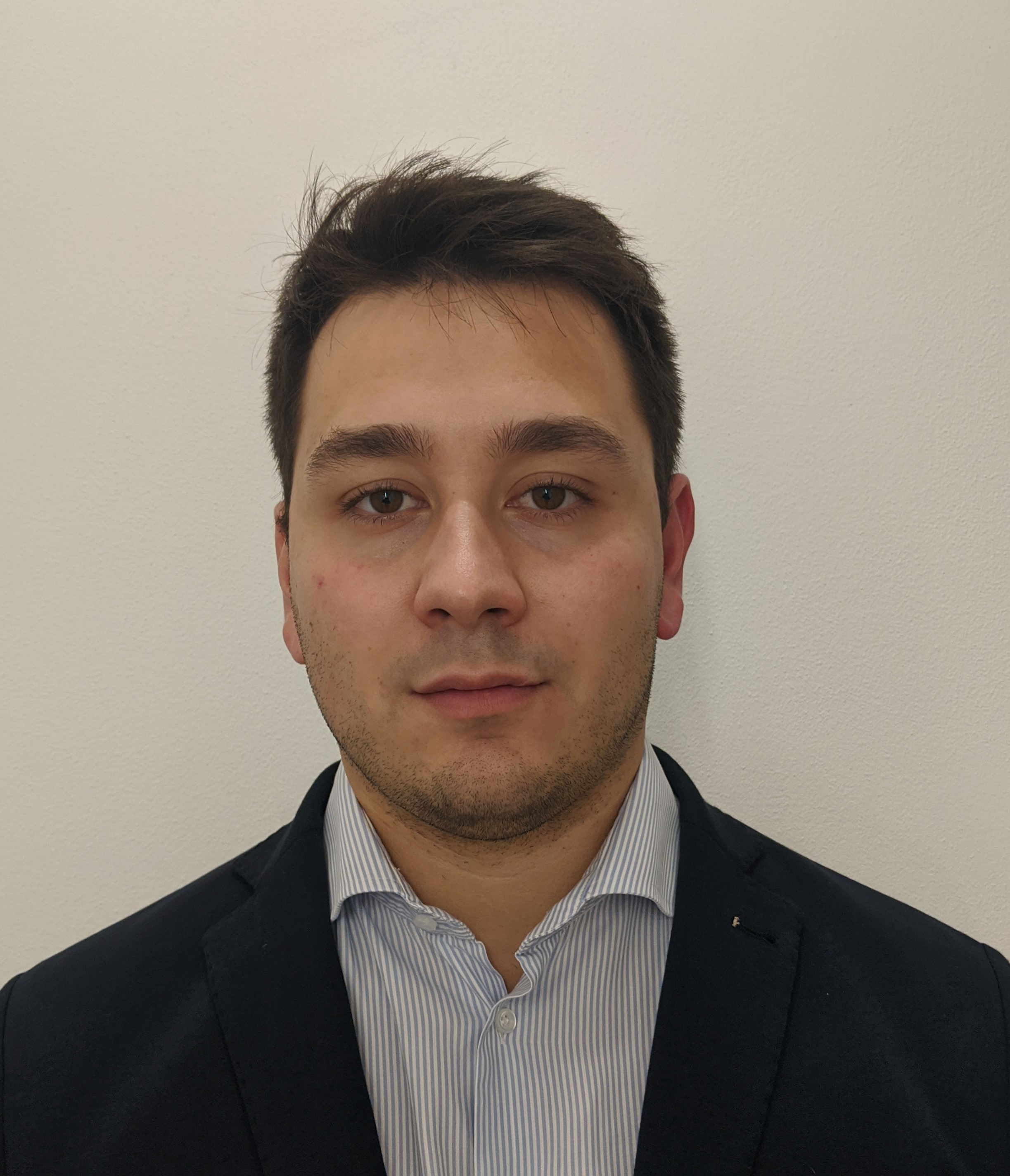}}]{Alberto Dalla Libera}
Alberto Dalla Libera received a Laurea degree in Control Engineering and the Ph.D. degree in information engineering from the University of Padua, Padua, Italy, in 2015 and 2019, respectively. 

He is currently a research fellow at the Department of Information Engineering of the University of Padua. His research interests include Robotics, Reinforcement Learning, Machine Learning, and Identification. In particular, he is interested in the application of Machine Learning techniques for modeling and control of physical systems.
\end{IEEEbiography}

\vfill
\end{document}